\pdfoutput=1  
\documentclass[10pt,journal]{IEEEtran}

\usepackage{amsmath,amssymb,amsfonts,mathtools}
\usepackage{graphicx}
\usepackage{booktabs}
\usepackage{multirow}
\usepackage{array}
\usepackage{threeparttable}
\usepackage{makecell}
\usepackage{xcolor}
\usepackage{algorithm}
\usepackage{algpseudocode}
\usepackage{tikz}
\usetikzlibrary{arrows.meta,positioning,calc,fit,backgrounds,shapes.geometric}
\usepackage{url}
\usepackage[hidelinks,breaklinks=true]{hyperref}
\usepackage{cite}

\newcommand{\vect}[1]{\boldsymbol{#1}}
\newcommand{\mat}[1]{\mathbf{#1}}
\newcommand{\real}{\mathbb{R}}
\DeclareMathOperator*{\argmax}{arg\,max}
\newcommand{\best}[1]{\textbf{#1}}
\newcommand{\yes}{\checkmark}

\newcolumntype{L}[1]{>{\raggedright\arraybackslash}p{#1}}
\newcolumntype{C}[1]{>{\centering\arraybackslash}p{#1}}

\begin{document}

\title{PairCLIP-SWA: Similarity-Guided Balancing and\\
Stochastic Weight Averaging for Generalized\\
Deepfake Image Detection with OpenCLIP}

\author{Md.~Nakib,
        Sudipto~Sarkar,
        A.~J.~A.~Jubayer~Talukder,
        and~Mohammad~Ariful~Haque\\[4pt]
\normalsize Department of Electrical and Electronic Engineering,\\
\normalsize BUET, Dhaka 1205, Bangladesh
\thanks{Manuscript prepared September 2026.}}

\IEEEtitleabstractindextext{%
\begin{abstract}
Deepfake detectors are almost always deployed as the single checkpoint that
scored best on a validation split, yet the reliability of that choice is rarely
measured. We measure it with PairCLIP-SWA, a single-backbone detector that
adapts an OpenCLIP ViT-L/14 visual tower by progressive block unfreezing on a
class-balanced corpus built by CLIP-similarity-guided one-to-one subsampling,
and averages the fine-tuning trajectory in weight space. Trained in-domain on
the IEEE SP Cup 2025 (DF-Wild-Cup) benchmark under the same protocol as the
strongest published three-backbone ensemble, the weight-averaged detector
reaches 97.04\% accuracy, 98.64\% AU-ROC and a 4.07\% equal error rate, against
94.46\%, 97.60\% and 5.00\%. Its training log shows why averaging matters: over
the averaged epochs the non-averaged checkpoint swings across 5.34 points of
validation accuracy while its validation loss more than doubles, whereas the
average is never worse and is up to 5.99 points better. On an
OpenForensics-derived benchmark the same instability becomes a misranking: the
checkpoint that validation accuracy selects (98.43\%) scores 88.85\% on a
disjoint 10{,}905-image test partition, rejecting 21.1\% of genuine photographs,
while the terminal average, ranked last on validation, scores 94.76\%. A study
of twenty-two in-house configurations on the identical DF-Wild-Cup split,
including preprocessing ablations, shows that CLIP pre-training rather than
ensembling drives accuracy, that secondary face cropping costs about 34 points,
and that frequency-domain inputs transfer poorly. We recommend per-class
selection criteria measured off the early-stopping split.
\end{abstract}

\begin{IEEEkeywords}
Deepfake detection, model selection, stochastic weight averaging, CLIP,
OpenCLIP, vision transformer, preprocessing, class imbalance, evaluation
methodology, media forensics.
\end{IEEEkeywords}}

\maketitle
\IEEEdisplaynontitleabstractindextext
\IEEEpeerreviewmaketitle
\pagestyle{plain}
\thispagestyle{plain}

\section{Introduction}
\IEEEPARstart{G}{enerative} adversarial
networks~\cite{goodfellow2014gan,karras2020stylegan2} and latent diffusion
models~\cite{rombach2022ldm} have made photorealistic face manipulation a
commodity, with well-documented consequences for fraud, non-consensual imagery
and disinformation~\cite{verdoliva2020media,tolosana2020survey}. Detection has
become a safety-critical signal-processing problem, and its central difficulty
is generalization: detectors that are near-saturated inside one manipulation
family degrade sharply when the generator, the compression pipeline or the
acquisition conditions change~\cite{wang2020cnngenerated,corvi2023diffusion}.

The IEEE Signal Processing Cup 2025 (DF-Wild-Cup) corpus was assembled to stress
exactly this regime. It merges eight public benchmarks under the DeepfakeBench
protocol~\cite{yan2023deepfakebench}, removes provenance metadata, and has a
$5{:}1$ fake-to-real imbalance. The strongest published result on it,
CAE-Net~\cite{bhattacharjee2026caenet}, reaches $94.46\%$ by fusing
EfficientNet-B0~\cite{tan2019efficientnet}, DeiT~\cite{touvron2021deit} and a
wavelet-fed ConvNeXt~\cite{liu2022convnext}, trained in five disjoint-subset
stages. Like many forensic detectors, CAE-Net is reported as a configuration
tuned on the very split it is evaluated on.

This paper started as a detector paper and became, in the course of the
experiments, a measurement paper about that last habit. We built a single
OpenCLIP ViT-L/14 detector that outperforms CAE-Net on every metric both
systems report. Its training log, however, showed the non-averaged checkpoint
oscillating by more than five points of validation accuracy from one epoch to
the next, and a second run on an OpenForensics-derived benchmark showed the
checkpoint that validation accuracy selected making twice as many errors on a
held-out partition as the checkpoint it ranked last. We report the detector,
but we organize the paper around what it revealed.

\subsection{Research Gaps}
\textbf{G1: Checkpoint selection is treated as free.} Forensic detectors are
reported and deployed as the best checkpoint on a validation split, often the
same split on which fusion weights or thresholds are tuned. Selection on a
finite validation set is itself a source of optimistic
bias~\cite{cawley2010overfitting,varma2006bias}; accuracy can fall
substantially on a freshly drawn sample of nominally the same
distribution~\cite{recht2019imagenet}; and in domain generalization the
selection protocol can matter as much as the method~\cite{gulrajani2021lostdg}.
We are not aware of a deepfake study that quantifies how unstable the selected
checkpoint is, or whether it is the right one to deploy.

\textbf{G2: Narrow pre-training priors.} Nearly all detectors reported on
DF-Wild-Cup initialize from ImageNet-supervised
weights~\cite{deng2009imagenet}. Contrastive image--text pre-training at web
scale~\cite{radford2021clip,cherti2023openclip,schuhmann2022laion5b} yields
features that separate synthetic from natural imagery across unseen
generators~\cite{ojha2023universal,cozzolino2024raising,khan2024clipping}, but
frozen probes under-fit fine-grained forensic evidence while unconstrained
fine-tuning distorts the pre-trained geometry~\cite{kumar2022finetuning}. How
much to adapt, and whether a single adapted CLIP tower can replace a
heterogeneous ensemble, are open questions.

\textbf{G3: Preprocessing choices are rarely justified or ablated.} Face
cropping, frequency transforms, augmentation and class balancing are usually
adopted by convention. Several of these choices can remove the very evidence a
detector needs --- blending boundaries~\cite{li2020facexray,shiohara2022sbi},
high-frequency residues~\cite{durall2020upconvolution} --- or open shortcuts
through class-conditional content such as identity~\cite{dong2023identityleakage}.

\subsection{Contributions}
\begin{enumerate}
\item \textbf{A measurement of checkpoint-selection reliability.} From the
complete training log we show that the non-averaged checkpoint swings across
$5.34$ points of DF-Wild-Cup validation accuracy while its validation loss
rises by a factor of $2.15$, and that the weight average is never worse and up
to $5.99$ points better (Section~\ref{sec:dynamics}). On a disjoint
$10{,}905$-image OpenForensics test partition, the validation-selected
checkpoint makes $2.1\times$ as many errors as the terminal average, through a
one-sided collapse of real-class recall that scalar accuracy on a balanced
split cannot detect (Section~\ref{sec:test}).

\item \textbf{A single-backbone detector that exceeds the best published
ensemble.} Trained in-domain under the same protocol, PairCLIP-SWA improves on
CAE-Net in accuracy ($97.04$ vs.\ $94.46\%$), AU-ROC ($98.64$ vs.\ $97.60\%$),
equal error rate ($4.07$ vs.\ $5.00\%$) and $F_1$, while training on $32.6\%$ of
the available training images in a single stage (Section~\ref{sec:spcup}).

\item \textbf{A justified, ablated preprocessing pipeline.} We state each
preprocessing and data-preparation decision together with its rationale, and
support them with ablations on the identical split: secondary face cropping,
FFT and wavelet inputs, class-balancing strategy, augmentation and training
objective (Sections~\ref{sec:preproc} and~\ref{sec:ablation}).

\item \textbf{A twenty-two-configuration design-space study}, including
negative results, showing that CLIP pre-training, not ensembling, is the
dominant factor, and that a weight-averaged single model outperforms a
five-model prediction ensemble with one forward pass instead of five
(Section~\ref{sec:ladder}).

\item \textbf{Evidence that cluster separability is not decision quality.} On
two corpora, t-SNE separation scores move independently of --- or opposite to
--- classifier accuracy, and an unsupervised partition of the $768$-dimensional
feature space shows only weak cluster structure (Section~\ref{sec:cluster}).
\end{enumerate}

Section~\ref{sec:related} reviews related work; Section~\ref{sec:method}
presents the preprocessing pipeline and the method;
Section~\ref{sec:exp} reports experiments; Section~\ref{sec:discussion}
gives ablations, the design-space study, discussion and limitations; and
Section~\ref{sec:conclusion} concludes.

\section{Related Work}\label{sec:related}

\subsection{Spatial, Frequency and Hybrid Detection}
CNN classifiers such as Xception~\cite{chollet2017xception} and
EfficientNet~\cite{tan2019efficientnet} remain strong single-model baselines on
FaceForensics++~\cite{rossler2019faceforensics} and
Celeb-DF~\cite{li2020celebdf}, refined by attention
modules~\cite{hu2018senet,woo2018cbam,zhao2021multiattentional}.
Up-sampling in generative decoders leaves spectral
signatures~\cite{durall2020upconvolution}, motivating detectors on the
DCT~\cite{frank2020frequency,ricker2024dct}, learned frequency
bands~\cite{qian2020f3net} and wavelet packets~\cite{wolter2022waveletpacket};
these cues are sharp but attenuated by re-compression and resizing. Face
X-ray~\cite{li2020facexray} and self-blended images~\cite{shiohara2022sbi}
instead target the blending boundary, arguably the most generator-agnostic
trace. CAE-Net~\cite{bhattacharjee2026caenet} fuses spatial, attention and
wavelet experts and addresses imbalance by splitting the $219{,}470$ DF-Wild-Cup
fakes into five disjoint subsets trained sequentially with all $42{,}690$ reals.

\subsection{CLIP for Forensics}
CLIP~\cite{radford2021clip} and OpenCLIP~\cite{ilharco2021openclip,cherti2023openclip}
learn joint image--text embeddings from web-scale pairs. Ojha \emph{et
al.}~\cite{ojha2023universal} showed that frozen CLIP features with a nearest
neighbour or linear classifier transfer across generators far better than
fine-tuned CNNs; Cozzolino \emph{et al.}~\cite{cozzolino2024raising} and Khan
and Dang-Nguyen~\cite{khan2024clipping} studied calibration and adaptation
strategies. Our progressive unfreezing, which descends from gradual unfreezing
in NLP~\cite{howard2018ulmfit}, sits between the frozen and fully fine-tuned
extremes, as the analysis of Kumar \emph{et al.}~\cite{kumar2022finetuning}
recommends. Neither component is new; our interest is in how such a model
behaves under selection.

\subsection{Shortcuts and Data Construction}
Detectors can exploit identity rather than manipulation
traces~\cite{dong2023identityleakage}, and self-blended images remove that
route by construction, at the price of biasing fakes towards one synthesis
operator~\cite{shiohara2022sbi}. Count-based balancing --- oversampling,
SMOTE~\cite{chawla2002smote}, loss reweighting, or partitioning --- leaves such
content differences untouched. Our subsampling step selects fakes by similarity
to reals in CLIP space, formally a greedy approximation to an assignment
problem~\cite{kuhn1955hungarian,burkard2012assignment}.

\subsection{Weight Averaging and Model Selection}
Stochastic weight averaging (SWA)~\cite{izmailov2018swa} averages iterates along
a trajectory and finds wider optima; model soups~\cite{wortsman2022soups}
average independently fine-tuned models. Selection on a validation set incurs
an optimistic bias that grows with the number of candidates
compared~\cite{cawley2010overfitting,varma2006bias}, and re-collected test sets
reveal gaps that validation accuracy did not
predict~\cite{recht2019imagenet}. Gulrajani and
Lopez-Paz~\cite{gulrajani2021lostdg} showed that under a fixed selection
protocol many domain-generalization methods fail to beat empirical risk
minimization. SWA itself is off-the-shelf; the contribution here is empirical
--- measuring, in a forensic setting, what the trajectory looks like and what
selecting from it costs.

\section{Preprocessing and Method}\label{sec:method}

\begin{figure*}[!t]
\centering
\resizebox{0.99\textwidth}{!}{%
\begin{tikzpicture}[
  font=\footnotesize,
  nbox/.style ={draw=black!55,rounded corners=2pt,align=center,inner sep=4pt,
                minimum height=13mm,minimum width=25mm,fill=white},
  ndata/.style={nbox,fill=black!6},
  nproc/.style={nbox,fill=blue!7,draw=blue!45},
  nres/.style ={nbox,fill=orange!12,draw=orange!60},
  narr/.style ={-{Latex[length=2.2mm]},draw=black!60,line width=0.6pt},
  nlbl/.style ={font=\scriptsize\itshape,text=black!55}
]
\node[ndata] (real)  at (0,0.95)  {Real pool $\mathcal{R}$\\$n_r=42{,}690$};
\node[ndata] (fake)  at (0,-0.95) {Fake pool $\mathcal{F}$\\$n_f=219{,}470$};
\node[nproc] (clip)  at (3.7,0)   {Frozen CLIP ViT-L/14\\$\phi(\cdot)\in\real^{768}$};
\node[nproc] (sim)   at (7.3,0)   {Cosine similarity\\$\mat{S}=\hat{\mat{U}}\hat{\mat{V}}^{\!\top}$};
\node[nproc] (match) at (10.9,0)  {Greedy one-to-one\\selection, no reuse};
\node[ndata] (corpus)at (14.7,0)  {Balanced corpus $\mathcal{D}_{\mathrm{pair}}$\\$85{,}380$ images, $1{:}1$};
\draw[narr] (real.east)  -- ++(0.55,0) |- (clip.west);
\draw[narr] (fake.east)  -- ++(0.55,0) |- (clip.west);
\draw[narr] (clip)  -- (sim);
\draw[narr] (sim)   -- (match);
\draw[narr] (match) -- (corpus);

\node[nproc] (sampler) at (14.7,-3.4) {Class-exact\\interleaved sampler};
\node[nproc] (aug)     at (11.0,-3.4) {Full frame, resize $224^2$\\H-flip, CLIP normalization};
\node[nproc] (tower)   at (7.0,-3.4)  {OpenCLIP ViT-L/14 tower\\$k(e)=\min(2{+}\lfloor e/2\rfloor,14)$};
\node[nproc] (head)    at (3.3,-3.4)  {Linear head\\$768\to2$, cross-entropy};
\node[nres]  (swa)     at (0,-3.4)    {SWA running average\\$\theta_{\mathrm{SWA}}=\frac{1}{n}\sum_t\theta_t$};
\draw[narr] (corpus)  -- (sampler);
\draw[narr] (sampler) -- (aug);
\draw[narr] (aug)     -- (tower);
\draw[narr] (tower)   -- (head);
\draw[narr] (head)    -- (swa);
\node[nlbl] at (7.0,-4.55) {frozen prefix: $\geq 41.98\%$ by design, $54.42\%$ in the reported run};

\node[nres] (eval)    at (0,-6.6)    {SP-Cup val.;\\OpenForensics val./test};
\node[nres] (sweep)   at (4.3,-6.6)  {Threshold sweep:\\AU-ROC, EER, $\tau^\star$};
\node[nres] (tsne)    at (8.6,-6.6)  {t-SNE, separation $S$,\\k-NN probe, DBSCAN};
\draw[narr] (swa)   -- (eval);
\draw[narr] (eval)  -- (sweep);
\draw[narr] (sweep) -- (tsne);

\begin{scope}[on background layer]
\node[fit=(real)(fake)(corpus),draw=black!30,dashed,rounded corners=3pt,
      inner sep=8pt,
      label={[font=\scriptsize\bfseries,text=black!65,yshift=-1mm]above left:%
      Stage 1: class balancing by CLIP-similarity-guided subsampling (P6)}] {};
\node[fit=(swa)(sampler),draw=black!30,dashed,rounded corners=3pt,
      inner sep=8pt,
      label={[font=\scriptsize\bfseries,text=black!65,yshift=-1mm]above left:%
      Stage 2: preprocessing (P1--P5, P7), partial adaptation, weight averaging}] {};
\node[fit=(eval)(tsne),draw=black!30,dashed,rounded corners=3pt,
      inner sep=8pt,
      label={[font=\scriptsize\bfseries,text=black!65,yshift=-1mm]above left:%
      Stage 3: evaluation, selection analysis and representation geometry}] {};
\end{scope}
\end{tikzpicture}}
\caption{Overview of PairCLIP-SWA and of the analyses in this paper. Labels
P1--P8 refer to the preprocessing steps of Table~\ref{tab:preproc}. Stage~1
builds a balanced training corpus from the DF-Wild-Cup training partition by
selecting, for each real image, its most similar unused fake in CLIP space.
Stage~2 applies trace-preserving preprocessing, adapts an OpenCLIP ViT-L/14
tower with a widening unfreezing schedule, and averages the trajectory. Stage~3
evaluates the resulting checkpoints and analyses selection and representation
geometry.}
\label{fig:pipeline}
\end{figure*}
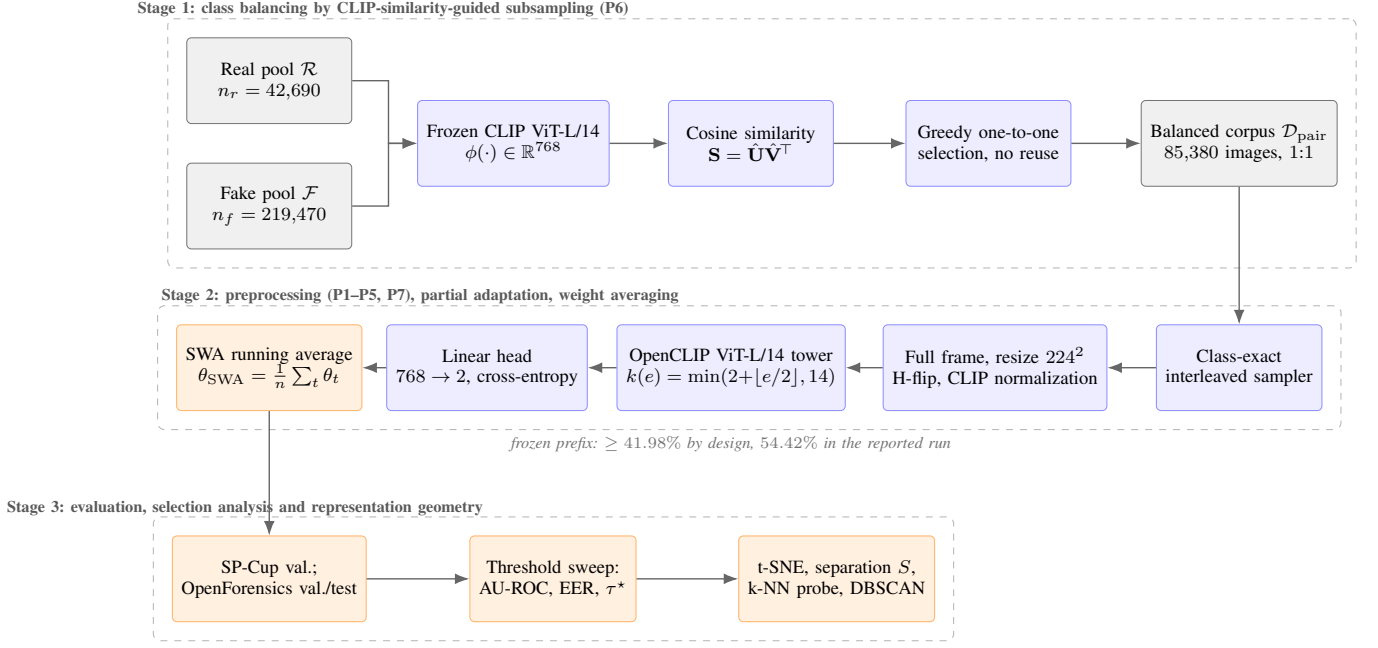

\subsection{Problem Statement and Notation}
Let $x$ be an RGB image and $y\in\{0,1\}$ its label. Our implementation uses
$y=1$ for \emph{real} and $y=0$ for \emph{fake}; all precision, recall and
$F_1$ values in this paper are nevertheless reported with the \emph{fake}
class positive, matching~\cite{bhattacharjee2026caenet}. A training corpus with
real pool $\mathcal{R}=\{x^r_i\}_{i=1}^{n_r}$ and fake pool
$\mathcal{F}=\{x^f_j\}_{j=1}^{n_f}$ yields a trajectory of parameters
$\theta_0,\theta_1,\dots$; a \emph{selection rule} maps the trajectory to one
deployed classifier $f_\theta$. We study both the classifier and the rule.

\subsection{Preprocessing Pipeline and Design Rationale}\label{sec:preproc}
Table~\ref{tab:preproc} lists every transformation between a stored image and
the network input, with the reason for each choice and the evidence that
supports it. Formally, with $R$ an anisotropic bilinear resampling to
$224\times224$, $F_b$ a horizontal flip applied when $b=1$, and $N$ the
per-channel standardization
\begin{equation}
N(\vect{v})_c=\frac{v_c/255-\mu_c}{\sigma_c},\qquad c\in\{\mathrm{R},\mathrm{G},\mathrm{B}\},
\label{eq:norm}
\end{equation}
with the CLIP statistics $\vect{\mu}=(0.4815,0.4578,0.4082)$ and
$\vect{\sigma}=(0.2686,0.2613,0.2758)$, the training and evaluation transforms
are
\begin{align}
T_{\mathrm{tr}}(x)&=N\big(F_b(R(x))\big),\qquad b\sim\mathrm{Bern}(0.5),
\label{eq:transforms}\\
T_{\mathrm{ev}}(x)&=N\big(R(x)\big).
\end{align}

\begin{table*}[!t]
\caption{Preprocessing and data-preparation pipeline. ``Evidence'' points to
the ablation or measurement that bears on each choice; ``design'' marks choices
motivated by argument only.}
\label{tab:preproc}
\centering
\footnotesize
\renewcommand{\arraystretch}{1.12}
\begin{tabular}{@{}lL{2.55cm}L{3.1cm}L{6.9cm}L{3.3cm}@{}}
\toprule
& Step & Setting & Rationale & Evidence \\
\midrule
P1 & Input region & Image as released; no secondary face detection or crop &
Blending boundaries and face--context inconsistencies are among the most
generator-agnostic cues~\cite{li2020facexray,shiohara2022sbi}; a tight crop
removes them. DF-Wild-Cup images are already aligned by
DeepfakeBench~\cite{yan2023deepfakebench}, so a second detector pass compounds
rather than removes variation, and fails silently on profile, occluded or
low-resolution faces. &
dlib crop: $56.05\%$ val., $\approx34$ points below full frame
(Table~\ref{tab:ablation}) \\
P2 & Decoding & 3-channel RGB (greyscale and alpha converted) &
Matches the input contract of CLIP pre-training; greyscale captures occur in
the wild (Fig.~\ref{fig:qual}). & design \\
P3 & Resampling & Bilinear resize to $224\times224$, no crop &
Reproduces the pre-training token grid ($16\times16$ patches of $14$ pixels) so
positional embeddings need no interpolation; resizing rather than cropping keeps
peripheral boundary evidence; the aspect distortion is identical for both
classes and cannot act as a label cue. & design \\
P4 & Normalization & Eq.~\eqref{eq:norm}, CLIP $\vect{\mu},\vect{\sigma}$ &
The frozen patch embedding and at least $41.98\%$ of the tower cannot adapt to
a different input scale; ImageNet statistics would shift the distribution the
frozen prefix sees. & Section~\ref{sec:repro}: a double rescale compresses
inputs $255\times$ \\
P5 & Augmentation & Horizontal flip, $p=0.5$, training only &
Flipping is label-preserving and leaves pixel statistics untouched; colour
jitter, blur, JPEG re-compression and random crops alter exactly the
high-frequency and colour evidence of manipulation, or remove peripheral
boundaries (P1, P3). & ViT+LoRA: augmentation raises the early peak but the gain
is not retained (Table~\ref{tab:ablation}) \\
P6 & Class balancing & One-to-one CLIP-similarity-guided subsample of the fake
pool, without replacement (Section~\ref{sec:pairing}) &
One balanced corpus in one training stage, instead of oversampling (repeated
reals), reweighting (majority domination), SMOTE (synthetic ``reals'' are
invalid for authenticity) or five sequential stages~\cite{bhattacharjee2026caenet};
intended to reduce class-conditional content differences~\cite{dong2023identityleakage}. &
Random undersampling, same backbone, different schedule: $93.36\%$
(Table~\ref{tab:ablation}) \\
P7 & Batch composition & Interleaved class-exact sampler, batch $32$ &
Removes batch-level label variance (standard deviation $8.8$ points of real
fraction under random batching); interleaving keeps each DataParallel shard
exactly balanced. & design \\
P8 & Evaluation & $T_{\mathrm{ev}}$; no flip, no test-time augmentation &
Deterministic predictions, so that checkpoint comparisons reflect weights
only. & design \\
\bottomrule
\end{tabular}
\end{table*}

Three of these choices deserve emphasis because they depart from common
practice. \emph{P1 and P3} keep the full released frame: whatever crop is
applied upstream, we do not crop again, because the evidence most likely to
transfer across generators lives at the boundary between the manipulated
region and its surroundings. \emph{P5} is deliberately minimal: in forensics,
photometric augmentation is not neutral, since blur, noise and compression are
themselves forgery-like transformations. We return in
Section~\ref{sec:discussion} to the cost of this choice --- a detector that has
never seen degraded \emph{real} images can learn that degradation implies
synthesis. \emph{P6} replaces multistage partitioning by a single subsampled
corpus; its motivation and its limits are discussed next.

\subsection{Stage 1: CLIP-Similarity-Guided Subsampling}\label{sec:pairing}
\subsubsection{Motivation}
Decompose an image into a content factor $c$ (identity, pose, illumination,
background, capture pipeline) and a manipulation factor $m$. If
$p(c\mid y{=}1)\neq p(c\mid y{=}0)$ in the training corpus, empirical risk
minimization can lower the loss using $c$ alone, a route that does not
transfer~\cite{dong2023identityleakage}. When the majority class must be
subsampled anyway, it is preferable to keep the fakes that most resemble the
reals rather than a random fifth of them.

\subsubsection{Objective and greedy solution}
With a frozen CLIP ViT-L/14 image encoder $\phi$ and row-normalized embeddings
$\hat{\mat{U}}\in\real^{n_r\times768}$, $\hat{\mat{V}}\in\real^{n_f\times768}$,
the similarity matrix and the ideal injective assignment are
\begin{equation}
\mat{S}=\hat{\mat{U}}\hat{\mat{V}}^{\!\top},\qquad
\pi^\star=\argmax_{\pi\ \text{injective}}\ \sum_{i=1}^{n_r}S_{i,\pi(i)},
\label{eq:assignment}
\end{equation}
a linear assignment problem~\cite{kuhn1955hungarian,burkard2012assignment}
whose exact solution costs $O(n^3)$. We solve it greedily, processing reals in
order and excluding fakes already used:
\begin{equation}
\pi(i)=\argmax_{j\notin\mathcal{U}_{i-1}} S_{ij},\qquad
\mathcal{U}_i=\mathcal{U}_{i-1}\cup\{\pi(i)\},\quad\mathcal{U}_0=\varnothing .
\label{eq:greedy}
\end{equation}
The no-reuse constraint prevents a few generic fakes from absorbing many real
partners. Rows are streamed in blocks of $B=1000$, so peak memory is
$O(Bn_f)$ --- one $1000\times219{,}470$ block occupies $837$\,MB in
\texttt{float32} --- while total work stays $O(n_rn_fd)$ for
$9.37\times10^{9}$ similarities. Algorithm~\ref{alg:pairing} states the
procedure.

\begin{algorithm}[!t]
\caption{CLIP-Similarity-Guided One-to-One Subsampling}
\label{alg:pairing}
\begin{algorithmic}[1]
\Require real pool $\mathcal{R}$, fake pool $\mathcal{F}$, frozen encoder
         $\phi$, block size $B$
\Ensure  balanced corpus $\mathcal{D}_{\mathrm{pair}}$
\State $\hat{\mat{U}}\gets \mathrm{rownorm}\big([\phi(x^r_i)]_i\big)$;\quad
       $\hat{\mat{V}}\gets \mathrm{rownorm}\big([\phi(x^f_j)]_j\big)$
\State $\mathcal{U}\gets\varnothing$;\quad $\mathcal{D}_{\mathrm{pair}}\gets\varnothing$
\For{each row block $b=1,\dots,\lceil n_r/B\rceil$}
    \State $\mat{S}^{(b)}\gets \hat{\mat{U}}^{(b)}\hat{\mat{V}}^{\!\top}$
           \Comment{$B\times n_f$}
    \For{each row $i$ of block $b$}
        \State $\pi(i)\gets\argmax_{j\notin\mathcal{U}} S^{(b)}_{ij}$;\quad
               $\mathcal{U}\gets\mathcal{U}\cup\{\pi(i)\}$
        \State $\mathcal{D}_{\mathrm{pair}}\gets\mathcal{D}_{\mathrm{pair}}
                \cup\{(x^r_i,\,x^f_{\pi(i)})\}$
    \EndFor
    \State free $\mat{S}^{(b)}$
\EndFor
\State \Return $\mathcal{D}_{\mathrm{pair}}$
\end{algorithmic}
\end{algorithm}

\subsubsection{Data usage}
Applied to the DF-Wild-Cup training partition ($n_r=42{,}690$,
$n_f=219{,}470$), Algorithm~\ref{alg:pairing} returns $42{,}690$ pairs, a
$1{:}1$ corpus of $85{,}380$ images: $32.6\%$ of the available training images,
keeping $19.5\%$ of the fakes and discarding $176{,}780$. By comparison, one
stage of CAE-Net's schedule already uses $86{,}584$ images and the full
schedule visits all $262{,}160$ five times~\cite{bhattacharjee2026caenet}. The
subsample is therefore also a data-efficiency choice, and it has a cost: most
of the generator diversity in the fake pool never reaches the detector.

\subsubsection{Scope of the claim}
Two caveats bound what this stage can be credited with. First, we have no
controlled comparison against a random one-to-one subsample under an identical
schedule (Section~\ref{sec:ablation}). Second, the implementation that
produced the reported corpus computed $\phi$ on inputs that had been rescaled
twice (Section~\ref{sec:repro}), so the similarities that drove the selection
were computed on strongly attenuated images rather than on the semantic
descriptor that motivates the design. For the reported run, Stage~1 should
therefore be read as a deterministic similarity-guided one-to-one subsample
whose content alignment has not been established.

\subsection{Stage 2: Class-Exact Sampling}\label{sec:sampler}
Each epoch draws independent permutations $\sigma_r,\sigma_f$ of the real and
fake indices of $\mathcal{D}_{\mathrm{pair}}$, interleaves them as
$\mathcal{L}=(\sigma_r(1),\sigma_f(1),\sigma_r(2),\sigma_f(2),\dots)$, truncates
$\mathcal{L}$ to a multiple of the batch size $32$, and cuts it into
contiguous batches. Every batch holds exactly $16$ real and $16$ fake images,
and because the order is interleaved, each of the two halves that
\texttt{nn.DataParallel} scatters to its replicas is also exactly balanced.
The ViT tower uses layer normalization~\cite{ba2016layernorm}, which is
per-sample, so the shard split does not change the forward computation.

\subsection{Stage 2: Backbone}\label{sec:arch}
The detector is the visual tower of OpenCLIP ViT-L/14~\cite{ilharco2021openclip}
with the original OpenAI weights~\cite{radford2021clip}, followed by a linear
head; the text tower is discarded. An input is split into $N_p=256$ patches of
$14\times14$ pixels, embedded to width $d_m=1024$, prefixed with a class token
and summed with positional embeddings:
$\mat{Z}_0=[\vect{z}_{\mathrm{cls}};\mathrm{E}\vect{x}_1;\dots;\mathrm{E}\vect{x}_{N_p}]
+\mat{E}_{\mathrm{pos}}$. Each of $L=24$ pre-norm
blocks~\cite{vaswani2017attention,dosovitskiy2021vit} applies
\begin{align}
\mat{Z}'_{\ell}&=\mat{Z}_{\ell-1}+\mathrm{MHSA}\big(\mathrm{LN}(\mat{Z}_{\ell-1})\big),\\
\mat{Z}_{\ell}&=\mat{Z}'_{\ell}+\mathrm{MLP}\big(\mathrm{LN}(\mat{Z}'_{\ell})\big),
\end{align}
with $16$-head scaled dot-product attention of head width $64$ and an
expansion-$4$ GELU MLP~\cite{hendrycks2016gelu}. The representation is the
projected class token and the classifier is affine:
\begin{align}
\vect{g}(x)&=\mathrm{LN}\big(\mat{Z}_L^{(0)}\big)\mat{W}_{\mathrm{proj}}\in\real^{768},
\label{eq:head}\\
f_\theta(x)&=\mathrm{softmax}\big(\mat{W}_c\vect{g}(x)+\vect{b}_c\big).
\end{align}
We keep $\mat{W}_{\mathrm{proj}}$ because the projected space is where CLIP's
contrastive geometry was learned and where frozen-feature transfer has been
demonstrated~\cite{ojha2023universal}. Table~\ref{tab:params} gives the
parameter budget.

\begin{table}[!tb]
\caption{Parameter budget and trainable fraction. The design cap $K_{\max}=14$
was not reached within the $20$-epoch budget of the reported run, which ended
at $k=11$ blocks.}
\label{tab:params}
\centering
\footnotesize
\begin{tabular}{@{}lrr@{}}
\toprule
Component & Parameters & Share \\
\midrule
Patch embedding ($3{\times}14{\times}14{\to}1024$) & $602{,}112$ & $0.20\%$ \\
Class token and positional embedding & $264{,}192$ & $0.09\%$ \\
Pre-/post-layer normalization & $4{,}096$ & $<\!0.01\%$ \\
Transformer blocks ($24\times12.60$\,M) & $302{,}309{,}376$ & $99.45\%$ \\
Output projection ($1024{\to}768$) & $786{,}432$ & $0.26\%$ \\
Linear head ($768{\to}2$) & $1{,}538$ & $<\!0.01\%$ \\
\midrule
\textbf{Total} & $\mathbf{303{,}967{,}746}$ & $100\%$ \\
\midrule
\multicolumn{3}{@{}l}{\emph{Trainable parameters}}\\
$k=2$ blocks (epoch $0$) & $25{,}193{,}986$ & $8.29\%$ \\
$k=8$ blocks (epochs $12$--$13$) & $100{,}771{,}330$ & $33.15\%$ \\
$k=11$ blocks (epochs $18$--$19$, reported run) & $138{,}560{,}002$ & $45.58\%$ \\
$k=14$ blocks (design cap) & $176{,}348{,}674$ & $58.02\%$ \\
\midrule
Frozen, reported run & $165{,}407{,}744$ & $54.42\%$ \\
Frozen, design guarantee & $127{,}619{,}072$ & $41.98\%$ \\
\bottomrule
\end{tabular}
\end{table}

\subsection{Stage 2: Progressive Unfreezing and Optimization}\label{sec:unfreeze}
At epoch $e$ the head and the last $k(e)$ blocks are trainable,
\begin{equation}
k(e)=\min\big(2+\lfloor e/2\rfloor,\;K_{\max}\big),\qquad K_{\max}=14,
\label{eq:schedule}
\end{equation}
and everything else, including the embeddings and $\mat{W}_{\mathrm{proj}}$,
stays frozen. At $e=0$ the model is a linear probe plus two blocks, so early
gradients act on an essentially intact CLIP manifold~\cite{kumar2022finetuning};
newly unfrozen tensors join AdamW as fresh parameter groups with zero moment
estimates. Training minimizes the unweighted two-way cross-entropy --- the
sampler already balances every batch --- with AdamW~\cite{loshchilov2019adamw}
at $\eta_0=10^{-4}$ and weight decay $\lambda=0$ (so the update is
Adam~\cite{kingma2015adam}), under \texttt{float16} automatic mixed
precision~\cite{micikevicius2018mixed} with dynamic loss scaling. Early
stopping monitors the non-averaged validation accuracy with patience $8$.

\subsection{Stage 2: Stochastic Weight Averaging}\label{sec:swa}
From epoch $e_{\mathrm{swa}}$ we maintain the running mean
\begin{equation}
\theta_{\mathrm{SWA}}^{(t)}=\theta_{\mathrm{SWA}}^{(t-1)}
  +\frac{\theta_t-\theta_{\mathrm{SWA}}^{(t-1)}}{n_t+1},
\label{eq:swa}
\end{equation}
the online form of $\frac{1}{n}\sum_t\theta_t$~\cite{izmailov2018swa}, and anneal
the learning rate towards $\eta_{\mathrm{swa}}=0.1\,\eta_0=10^{-5}$ with the
ten-epoch cosine schedule of \texttt{torch.optim.swa\_utils.SWALR}, stepped once
per epoch. Both the running average and the non-averaged iterate are evaluated
and checkpointed every epoch, giving three deployable models:
\emph{best regular}, the non-averaged iterate with the highest validation
accuracy; \emph{best SWA}, the running average truncated at its best
validation epoch; and \emph{final SWA}, the average over the whole averaged
trajectory. A normalization-statistics pass is applied to the final average;
it leaves a layer-normalized tower unchanged, but is required for backbones
with batch normalization~\cite{ioffe2015batchnorm}. Algorithm~\ref{alg:train}
summarizes training.

\begin{algorithm}[!t]
\caption{PairCLIP-SWA Training and Checkpointing}
\label{alg:train}
\begin{algorithmic}[1]
\Require corpus $\mathcal{D}_{\mathrm{pair}}$, validation set $\mathcal{V}$,
         epochs $E$, patience $P$, SWA start $e_{\mathrm{swa}}$
\State $\theta\gets$ CLIP weights $+$ random head;\
       $a^\star\gets\mathrm{Acc}(f_\theta,\mathcal{V})$;\
       $a^\star_{\mathrm{swa}}\gets 0$;\ $p\gets 0$
\For{$e=0,\dots,E$}
    \State unfreeze last $k(e)$ blocks (Eq.~\eqref{eq:schedule}); extend optimizer
    \For{each class-exact batch}
        \State AMP forward, scaled backward, AdamW step
    \EndFor
    \State $a\gets\mathrm{Acc}(f_\theta,\mathcal{V})$
    \If{$e\geq e_{\mathrm{swa}}$}
        \State update $\theta_{\mathrm{SWA}}$ (Eq.~\eqref{eq:swa}); step SWALR
        \State $a_{\mathrm{swa}}\gets\mathrm{Acc}(f_{\theta_{\mathrm{SWA}}},\mathcal{V})$
        \If{$a_{\mathrm{swa}}>a^\star_{\mathrm{swa}}$}
            \State $a^\star_{\mathrm{swa}}\gets a_{\mathrm{swa}}$; save best SWA
        \EndIf
    \EndIf
    \If{$a>a^\star$} \State $a^\star\gets a$; $p\gets0$; save best regular
    \Else \State $p\gets p+1$; \textbf{if} $p\geq P$ \textbf{then break}
    \EndIf
\EndFor
\State normalization pass on $\theta_{\mathrm{SWA}}$; save final SWA
\end{algorithmic}
\end{algorithm}

\subsection{Stage 3: Representation Geometry}\label{sec:cluster-method}
For a subsample of held-out images we project the features $\vect{g}(x)$ of
Eq.~\eqref{eq:head} to two dimensions with t-SNE~\cite{maaten2008tsne}. With
class centroids $\vect{\mu}_f,\vect{\mu}_r$ and root-mean-square radial
dispersions $\sigma_f,\sigma_r$, we report the inter-centroid distance
$\Delta=\lVert\vect{\mu}_f-\vect{\mu}_r\rVert_2$ and the scale-free ratio
\begin{equation}
S=\frac{\Delta}{\sigma_f+\sigma_r},
\label{eq:sep}
\end{equation}
a two-cluster analogue of a Fisher ratio. A k-nearest-neighbour classifier fit
in the embedding serves as a probe of how much class information survives the
projection. As an independent check that does not depend on a projection, we
also partition the $768$-dimensional features with
DBSCAN~\cite{ester1996dbscan} and report the silhouette
coefficient~\cite{rousseeuw1987silhouette} and Davies--Bouldin
index~\cite{davies1979index}.

\section{Experiments}\label{sec:exp}

\subsection{Datasets and Splits}
\textbf{SP-Cup 2025 (DF-Wild-Cup).} Eight benchmarks --- Celeb-DF v1 and
v2~\cite{li2020celebdf}, FaceForensics++~\cite{rossler2019faceforensics},
DeepfakeDetection, FaceShifter~\cite{li2019faceshifter},
UADFV~\cite{yang2019uadfv}, and the DFDC preview and full
releases~\cite{dolhansky2020dfdc} --- under DeepfakeBench
preprocessing~\cite{yan2023deepfakebench}. The training partition has
$42{,}690$ real and $219{,}470$ fake images; the official validation partition
has $1{,}548$ real and $1{,}524$ fake images. The detector is trained on the
Stage-1 subsample of the training partition and selected and evaluated on the
validation partition. CAE-Net~\cite{bhattacharjee2026caenet} is likewise trained
on the training partition and evaluated on the same $3{,}072$ images, with
fusion weights tuned on them, so the comparison is in-domain and like-for-like.

\textbf{OpenForensics-derived benchmark.} A public in-the-wild face corpus
derived from OpenForensics~\cite{le2021openforensics} with fixed partitions. A
second, independent run of the same backbone, schedule and averaging procedure
is trained on its training partition, selected on its validation partition
($39{,}428$ images; $19{,}641$ fake, $19{,}787$ real) and evaluated once on its
test partition ($10{,}905$ images; $5{,}492$ fake, $5{,}413$ real), which is
used for neither gradients nor selection. Table~\ref{tab:data} summarizes the
role of every split.

\begin{table}[!tb]
\caption{Splits and their roles. Only the OpenForensics test partition is
untouched by both training and selection.}
\label{tab:data}
\centering
\footnotesize
\setlength{\tabcolsep}{3pt}
\begin{tabular}{@{}llrcl@{}}
\toprule
Corpus & Split & Images & Real/Fake & Role \\
\midrule
\multirow{2}{*}{\makecell[l]{DF-Wild-\\Cup}} & train$^\ast$ & $85{,}380$ & $1{:}1$ & training \\
 & val. & $3{,}072$ & $1548/1524$ & selection, eval. \\
\midrule
\multirow{3}{*}{\makecell[l]{Open-\\Forensics}} & train & --- & --- & training \\
 & val. & $39{,}428$ & $19787/19641$ & selection \\
 & test & $10{,}905$ & $5413/5492$ & \textbf{held-out test} \\
\bottomrule
\multicolumn{5}{@{}l}{\scriptsize $^\ast$Stage-1 subsample of $262{,}160$ training images.}
\end{tabular}
\end{table}

\subsection{Implementation and Reproducibility}\label{sec:repro}
The reported DF-Wild-Cup run was trained on two NVIDIA T4 GPUs with
\texttt{nn.DataParallel} in PyTorch~\cite{paszke2019pytorch}; several earlier
exploratory runs of Section~\ref{sec:ladder} used a single Tesla P100.
Table~\ref{tab:hyper} lists the configuration. Training starts from the OpenAI
CLIP weights with a randomly initialized head, which scores $45.80\%$ on the
validation set before adaptation, and runs for epochs $0$--$19$. Averaging
starts at $e_{\mathrm{swa}}=12$; because \texttt{AveragedModel} copies rather
than averages on its first update, the final average comprises the seven
iterates of epochs $13$--$19$.

\begin{table}[!tb]
\caption{Configuration of the reported DF-Wild-Cup run.}
\label{tab:hyper}
\centering
\footnotesize
\begin{tabular}{@{}ll@{}}
\toprule
Setting & Value \\
\midrule
Backbone & OpenCLIP ViT-L/14, OpenAI weights \\
Initialization & pre-trained tower, random linear head \\
Preprocessing & Table~\ref{tab:preproc} (P1--P8) \\
Training corpus & $85{,}380$-image $1{:}1$ subsample \\
Batch size & $32$ ($16{+}16$), $2\times$ T4, \texttt{DataParallel} \\
Optimizer & AdamW, $\eta_0=10^{-4}$, $\lambda=0$ \\
Precision & AMP \texttt{float16}, dynamic loss scaling \\
Epochs & $0$--$19$ \\
Unfreezing & $k(e)=\min(2+\lfloor e/2\rfloor,14)$; reached $k=11$ \\
SWA & start epoch $12$; SWALR to $10^{-5}$, cosine, $10$ epochs \\
Early stopping & patience $8$ on non-averaged val.\ accuracy \\
Stage-1 encoder & CLIP ViT-L/14 (HF \texttt{transformers}), frozen \\
Stage-1 block size & $B=1000$ \\
\bottomrule
\end{tabular}
\end{table}

Three implementation details matter for reproduction.

\emph{Double rescaling in Stage~1.} The subsampling code loads CLIP ViT-L/14
through HuggingFace \texttt{transformers}~\cite{wolf2020transformers} --- the
same OpenAI weights as the training tower --- but applies torchvision's
\texttt{ToTensor}, which already maps intensities to $[0,1]$, before
\texttt{CLIPProcessor}, whose default \texttt{do\_rescale} divides by $255$
again. We reproduced this path with \texttt{transformers}~4.46.3: the processor
emits a warning and rescales, so that the normalized inputs become
$(v_c/255^2-\mu_c)/\sigma_c\in[-1.792,-1.466]$ instead of spanning roughly
$[-1.8,2.1]$. Each channel's dynamic range shrinks by $255\times$; a random
noise image and a flat grey image map to nearly identical tensors. The Stage-1
similarities of the reported run were therefore computed on near-black inputs,
which is why Section~\ref{sec:pairing} limits the claim made for this stage.
Passing PIL images, or setting \texttt{do\_rescale=False}, restores the intended
path.

\emph{QuickGELU.} Instantiating \texttt{ViT-L-14} with the \texttt{openai} tag in
\texttt{open\_clip} pairs weights trained with QuickGELU with a configuration
using standard GELU. Fine-tuning absorbs the mismatch, but frozen-feature
reproductions should be aware of it.

\emph{In-loop versus re-evaluated accuracy.} Validation accuracy logged inside
the training loop is computed under \texttt{float16} autocast; saved checkpoints
are re-evaluated in full precision. The two differ by up to $0.21$ points (for
example, best SWA: $97.17\%$ in-loop at epoch $18$, $97.04\%$ re-evaluated). All
comparisons use re-evaluated values; trajectory figures necessarily use in-loop
values.

\subsection{Evaluation Protocol}
We report accuracy with $95\%$ Wilson score intervals~\cite{wilson1927interval},
\begin{equation}
\mathrm{CI}_{95}=\frac{\hat{p}+\frac{z^2}{2n}\pm z\sqrt{\frac{\hat{p}(1-\hat{p})}{n}
      +\frac{z^2}{4n^2}}}{1+\frac{z^2}{n}},\quad \hat p=\frac{k}{n},\ z=1.96.
\end{equation}
For the best-SWA DF-Wild-Cup checkpoint, softmax scores were retained and swept
over $1{,}000$ thresholds, giving AU-ROC, equal error rate (EER), the
accuracy-optimal threshold $\tau^\star$ and the precision, recall and $F_1$ at
$\tau^\star$. That sweep logged precision, recall and $F_1$ with the \emph{real}
class positive; because the class counts are known, the confusion matrix at
$\tau^\star$ is uniquely recoverable from the logged values
($1{,}533$ of $1{,}548$ real and $1{,}456$ of $1{,}524$ fake images correct, the
only integer solution consistent with all four logged figures), and we report
fake-positive values derived from it. For the OpenForensics splits, full
confusion matrices were logged at $\tau=0.5$. We also report specificity and
the Matthews correlation coefficient (MCC)~\cite{matthews1975mcc}.

\subsection{Results Across Splits}
\begin{table}[!tb]
\caption{Accuracy (\%) of the three checkpoints on every evaluated split, at
$\tau=0.5$, with $95\%$ Wilson intervals. Best per split in bold.}
\label{tab:main}
\centering
\footnotesize
\setlength{\tabcolsep}{2.4pt}
\begin{tabular}{@{}lccc@{}}
\toprule
Split ($n$) & Best regular & Best SWA & Final SWA \\
\midrule
DF-Wild-Cup val.\ ($3{,}072$)
  & $95.67$ & $\best{97.04}$ & $\best{97.04}$ \\
\emph{95\% CI}
  & \emph{[94.89, 96.34]} & \emph{[96.38, 97.58]} & \emph{[96.38, 97.58]} \\[2pt]
OpenForensics val.\ ($39{,}428$)
  & $\best{98.43}$ & $\best{98.43}$ & $97.67$ \\
\emph{95\% CI}
  & \emph{[98.30, 98.55]} & \emph{[98.30, 98.55]} & \emph{[97.52, 97.81]} \\[2pt]
OpenForensics test ($10{,}905$)
  & $88.85$ & $88.85$ & $\best{94.76}$ \\
\emph{95\% CI}
  & \emph{[88.24, 89.43]} & \emph{[88.24, 89.43]} & \emph{[94.33, 95.17]} \\
\bottomrule
\end{tabular}
\end{table}

\begin{table*}[!t]
\caption{Training log of the reported DF-Wild-Cup run over the averaged epochs
(in-loop values). ``Gap'' is SWA minus non-averaged validation accuracy; the two
coincide at epochs $12$--$13$ because the first SWA update copies the iterate.
Over epochs $12$--$19$ the non-averaged validation accuracy spans $5.34$ points
(s.d.\ $2.13$) against $1.86$ (s.d.\ $0.70$) for the average, with a mean change
between consecutive epochs of $2.33$ vs.\ $0.34$ points. Accuracy before
adaptation is $45.80\%$.}
\label{tab:trajectory}
\centering
\footnotesize
\setlength{\tabcolsep}{6pt}
\begin{tabular}{@{}crrrrrrrrr@{}}
\toprule
& & & \multicolumn{2}{c}{Training} & \multicolumn{2}{c}{Non-averaged val.} &
\multicolumn{2}{c}{SWA val.} & \\
\cmidrule(lr){4-5}\cmidrule(lr){6-7}\cmidrule(lr){8-9}
Epoch & Blocks & Trainable & Loss & Acc. & Loss & Acc. & Loss & Acc. & Gap \\
\midrule
12 & 8  & $33.15\%$ & $0.1628$ & $93.58$ & $0.1677$ & $95.70$ & $0.1677$ & $95.70$ & --- \\
13 & 8  & $33.15\%$ & $0.0802$ & $97.01$ & $0.1684$ & $95.61$ & $0.1684$ & $95.61$ & --- \\
14 & 9  & $37.30\%$ & $0.1060$ & $95.93$ & $0.2552$ & $91.05$ & $0.1383$ & $95.61$ & $+4.56$ \\
15 & 9  & $37.30\%$ & $0.0669$ & $97.44$ & $0.2005$ & $93.88$ & $0.1426$ & $95.31$ & $+1.43$ \\
16 & 10 & $41.44\%$ & $0.0885$ & $96.65$ & $0.2344$ & $92.22$ & $0.1386$ & $95.80$ & $+3.58$ \\
17 & 10 & $41.44\%$ & $0.0538$ & $97.89$ & $\mathbf{0.3603}$ & $\mathbf{90.36}$ & $0.1334$ & $96.35$ & $\mathbf{+5.99}$ \\
18 & 11 & $45.58\%$ & $0.0697$ & $97.29$ & $0.2307$ & $93.49$ & $\mathbf{0.1178}$ & $\mathbf{97.17}$ & $+3.68$ \\
19 & 11 & $45.58\%$ & $0.0351$ & $98.48$ & $0.2226$ & $95.67$ & $0.1227$ & $97.04$ & $+1.37$ \\
\bottomrule
\end{tabular}
\end{table*}

\begin{figure*}[!t]
\centering
\includegraphics[width=\textwidth]{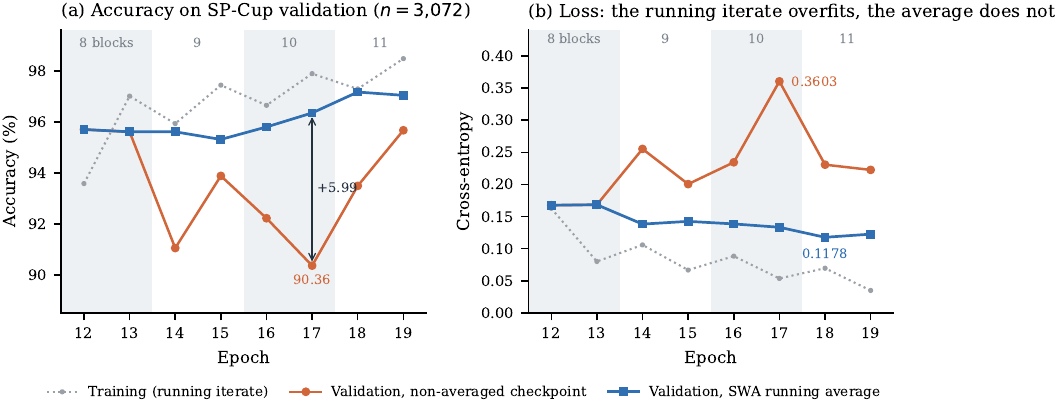}
\caption{Training dynamics of the reported DF-Wild-Cup run. (a) Validation
accuracy of the non-averaged checkpoint oscillates across $5.34$ points while
the weight average stays within $1.86$ points and is never below it; at epoch
$17$ the gap is $5.99$ points. (b) Training loss falls by $78\%$ while the
non-averaged validation loss rises by a factor of $2.15$ --- overfitting of the
adapted suffix --- whereas the averaged model's validation loss falls by $30\%$.
Shading marks the number of unfrozen blocks.}
\label{fig:trajectory}
\end{figure*}

Table~\ref{tab:main} already contains the paper's argument in compressed form.
On the two selection splits, the checkpoint that wins differs: averaging wins
by $1.37$ points on DF-Wild-Cup and loses by $0.76$ points on OpenForensics.
On the only split used for neither training nor selection, averaging wins by
$5.91$ points. Sections~\ref{sec:dynamics}--\ref{sec:test} explain both
patterns.

\begin{table}[!tb]
\caption{Accuracy on the DF-Wild-Cup validation set ($n=3{,}072$). Baselines as
reported in~\cite{bhattacharjee2026caenet}; ensemble rows are means over five
seeds. All systems are trained on the DF-Wild-Cup training partition.}
\label{tab:spcup}
\centering
\footnotesize
\begin{tabular}{@{}llrr@{}}
\toprule
Family & Model & Params & Acc.\ (\%) \\
\midrule
\multirow{7}{*}{Spatial CNN}
 & VGG-19~\cite{simonyan2015vgg}          & $144$\,M  & $62.25$ \\
 & ResNet-50~\cite{he2016resnet}          & $25.6$\,M & $69.00$ \\
 & MobileNet-V2                            & $3.5$\,M  & $78.12$ \\
 & ResNet-34~\cite{he2016resnet}          & $21.3$\,M & $80.32$ \\
 & Inception-V3                            & $26$\,M   & $84.38$ \\
 & ResNet-34 $+$ SE~\cite{hu2018senet}    & $21.3$\,M & $88.90$ \\
 & EfficientNet-B0~\cite{tan2019efficientnet} & $3.96$\,M & $90.79$ \\
\midrule
\multirow{4}{*}{Transformer}
 & ViT-B~\cite{dosovitskiy2021vit}        & $86.6$\,M & $78.12$ \\
 & Swin-T~\cite{liu2021swin}              & $28.3$\,M & $86.52$ \\
 & MobileViT                               & $5.6$\,M  & $87.04$ \\
 & DeiT-B~\cite{touvron2021deit}          & $85.6$\,M & $89.32$ \\
\midrule
\multirow{4}{*}{Frequency}
 & ResNet-34 $+$ FFT                       & $21.3$\,M & $66.55$ \\
 & ResNet-34 $+$ DCT~\cite{frank2020frequency} & $21.3$\,M & $71.48$ \\
 & Xception $+$ Wavelet                    & $20.8$\,M & $87.76$ \\
 & ConvNeXt $+$ Wavelet~\cite{liu2022convnext} & $27.8$\,M & $89.49$ \\
\midrule
\multirow{3}{*}{Ensemble}
 & EfficientNet $+$ DeiT                   & $89.6$\,M  & $92.59$ \\
 & EfficientNet $+$ ConvNeXt               & $31.8$\,M  & $93.14$ \\
 & \textbf{CAE-Net}~\cite{bhattacharjee2026caenet} & $117.4$\,M & $94.46$ \\
\midrule
\multirow{3}{*}{\makecell[l]{OpenCLIP\\(ours)}}
 & PairCLIP, best regular                  & $304.0$\,M & $95.67$ \\
 & \textbf{PairCLIP-SWA}, best/final SWA   & $304.0$\,M & $\best{97.04}$ \\
 & \quad at tuned threshold $\tau^\star$  & $304.0$\,M & $97.30^\dagger$ \\
\bottomrule
\multicolumn{4}{@{}l}{\scriptsize $^\dagger$Threshold selected on the same split; optimistic.}
\end{tabular}
\end{table}

\begin{table}[!tb]
\caption{Head-to-head on the DF-Wild-Cup validation set ($n=3{,}072$), fake
class positive. CAE-Net values as published~\cite{bhattacharjee2026caenet}.
PairCLIP-SWA is the best-SWA checkpoint.}
\label{tab:head2head}
\centering
\footnotesize
\setlength{\tabcolsep}{2.6pt}
\begin{threeparttable}
\begin{tabular}{@{}lccc@{}}
\toprule
Metric & CAE-Net & \multicolumn{2}{c}{PairCLIP-SWA} \\
\cmidrule(l){3-4}
 & & $\tau=0.5$ & $\tau^\star=0.680$\tnote{a} \\
\midrule
Accuracy (\%)             & $94.46$ & $\best{97.04}$ & $97.30$ \\
\emph{95\% CI}            & \emph{[93.60, 95.22]} & \emph{[96.38, 97.58]} & \emph{[96.66, 97.82]} \\
Errors / $3{,}072$         & $170$   & $\best{91}$ & $83$ \\
AU-ROC (\%)\tnote{b}      & $97.60$ & \multicolumn{2}{c}{$\best{98.64}$} \\
EER (\%)\tnote{b}         & $5.00$  & \multicolumn{2}{c}{$\best{4.07}$} \\
Precision (\%)            & $93.70$ & n/l & $98.98$ \\
Recall (\%)               & $95.60$ & n/l & $95.54$ \\
Specificity (\%)          & ---     & n/l & $99.03$ \\
$F_1$ (\%)                & $94.64$ & n/l & $97.23$ \\
MCC                       & ---     & n/l & $0.947$ \\
\midrule
Backbones / passes        & $3$ / $3$ & \multicolumn{2}{c}{$\best{1}$ / $\best{1}$} \\
Parameters                & $\best{117.4}$\,M & \multicolumn{2}{c}{$304.0$\,M} \\
GMACs\tnote{c}            & $\best{24.8}$ & \multicolumn{2}{c}{$81.0$} \\
ms / image\tnote{d}       & $10.4$ (P100) & \multicolumn{2}{c}{$28.1$ ($2\times$T4)} \\
Training images           & $262{,}160$ & \multicolumn{2}{c}{$\best{85{,}380}$} \\
Training stages           & $5$ & \multicolumn{2}{c}{$\best{1}$} \\
\bottomrule
\end{tabular}
\begin{tablenotes}[flushleft]\footnotesize
\item[a] Accuracy-optimal threshold on $P(\text{real})$, selected on this split;
optimistic. Precision/recall/$F_1$ at $\tau=0.5$ were not logged (n/l).
\item[b] Threshold-free.
\item[c] Ours computed analytically, Eq.~\eqref{eq:macs}.
\item[d] Different hardware; CAE-Net from its reported throughput of
$96.42$ images/s at batch $32$; ours is the average per-image time over the
validation pass.
\end{tablenotes}
\end{threeparttable}
\end{table}

\begin{figure}[!tb]
\centering
\includegraphics[width=\columnwidth]{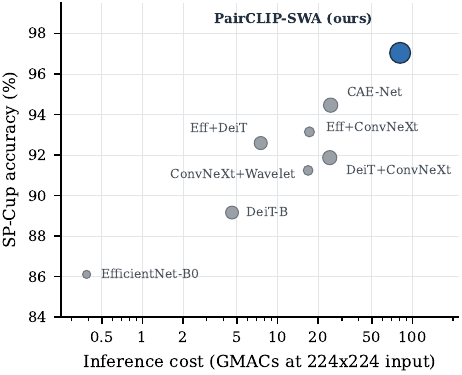}
\caption{Accuracy versus inference cost on DF-Wild-Cup; marker area is
proportional to parameter count. Grey markers are the single models and
ensembles of~\cite{bhattacharjee2026caenet}.}
\label{fig:tradeoff}
\end{figure}

\subsection{Training Dynamics: the Case for Averaging}\label{sec:dynamics}
Table~\ref{tab:trajectory} and Fig.~\ref{fig:trajectory} give the complete log
over the averaged epochs. Three observations follow.

\emph{The selected checkpoint is a noisy maximum.} Between consecutive epochs
the non-averaged validation accuracy moves by $2.33$ points on average, and
over eight epochs it spans $5.34$ points, from $95.70$ to $90.36\%$. Any
selection rule that picks one of these iterates is picking from a distribution
whose width is almost three times the $1.87$-point margin separating the three
best systems reported in~\cite{bhattacharjee2026caenet}.

\emph{The iterate overfits; the average does not.} Training loss falls from
$0.1628$ to $0.0351$ ($-78\%$) and training accuracy rises to $98.48\%$, while
the non-averaged validation loss climbs to $0.3603$ at epoch~$17$, $2.15$ times
its epoch-$12$ value, with a train--validation accuracy gap of $7.53$ points.
Over the same epochs the averaged model's validation loss falls to $0.1178$
($-30\%$) and its gap at epoch $19$ is $1.44$ points. The first large increase
in non-averaged validation loss (epoch $14$, $+0.087$) coincides with the first
unfreezing event inside the averaged window; the largest (epoch $17$,
$+0.126$) follows the second.

\emph{Averaging dominates at every epoch.} From epoch $14$, where the average
first differs from the iterate, the average is better at every epoch, by
$1.37$ to $5.99$ points ($3.44$ on average). Whatever epoch early stopping
might have landed on, the averaged model was the better deployable choice.
This is the in-sample half of the selection argument; the out-of-sample half
requires a split untouched by selection (Section~\ref{sec:test}).

\begin{table}[!tb]
\caption{Complete metrics on OpenForensics at $\tau=0.5$, fake class positive,
from the confusion matrices of Fig.~\ref{fig:cmof}.}
\label{tab:ofmetrics}
\centering
\footnotesize
\setlength{\tabcolsep}{2.8pt}
\begin{tabular}{@{}llrrrrrr@{}}
\toprule
Split & Checkpoint & Acc. & Prec. & Rec. & Spec. & $F_1$ & MCC \\
\midrule
\multirow{2}{*}{\makecell[l]{Validation\\($39{,}428$)}}
 & Best reg./SWA & $\best{98.43}$ & $97.84$ & $\best{99.03}$ & $97.83$ & $\best{98.43}$ & $\best{0.969}$ \\
 & Final SWA     & $97.67$ & $\best{99.58}$ & $95.73$ & $\best{99.60}$ & $97.61$ & $0.954$ \\
\midrule
\multirow{2}{*}{\makecell[l]{Test\\($10{,}905$)}}
 & Best reg./SWA & $88.85$ & $82.58$ & $\best{98.67}$ & $78.88$ & $89.91$ & $0.792$ \\
 & Final SWA     & $\best{94.76}$ & $\best{94.97}$ & $94.61$ & $\best{94.92}$ & $\best{94.79}$ & $\best{0.895}$ \\
\bottomrule
\end{tabular}
\end{table}

\begin{figure}[!tb]
\centering
\includegraphics[width=0.95\columnwidth]{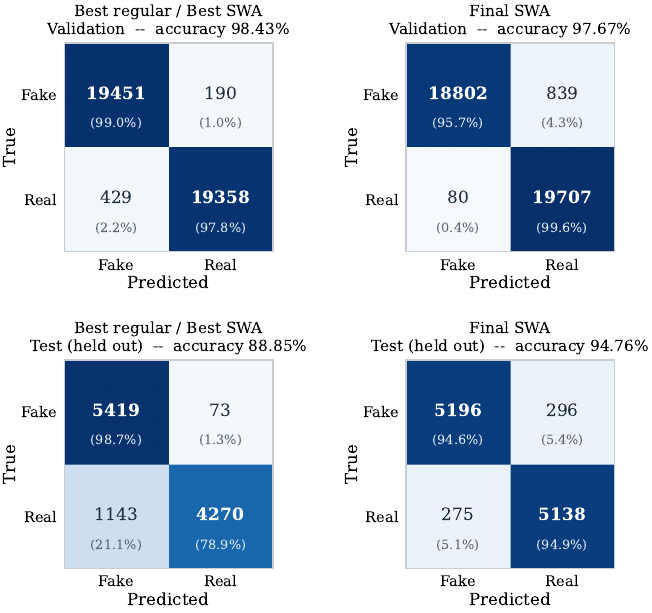}
\caption{OpenForensics confusion matrices for the validation-selected checkpoint
(left) and the terminal average (right), on validation (top) and held-out test
(bottom). Row percentages are per-class recall.}
\label{fig:cmof}
\end{figure}

\begin{figure*}[!t]
\centering
\includegraphics[width=\textwidth]{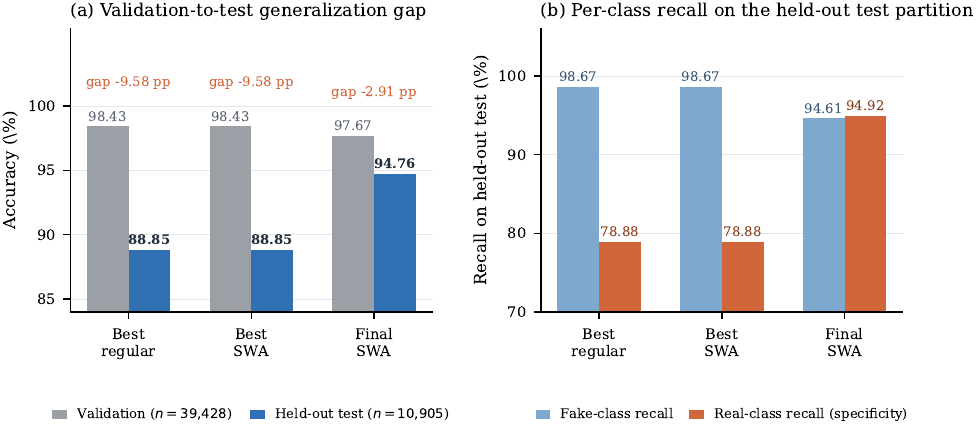}
\caption{(a) OpenForensics accuracy on validation (grey) and held-out test
(blue). The validation-selected checkpoints lose $9.58$ points in transfer; the
terminal average loses $2.91$. (b) Test recall per class: the entire loss of
the selected checkpoints is carried by the real class.}
\label{fig:valtest}
\end{figure*}

\begin{figure*}[!t]
\centering
\includegraphics[width=0.73\textwidth]{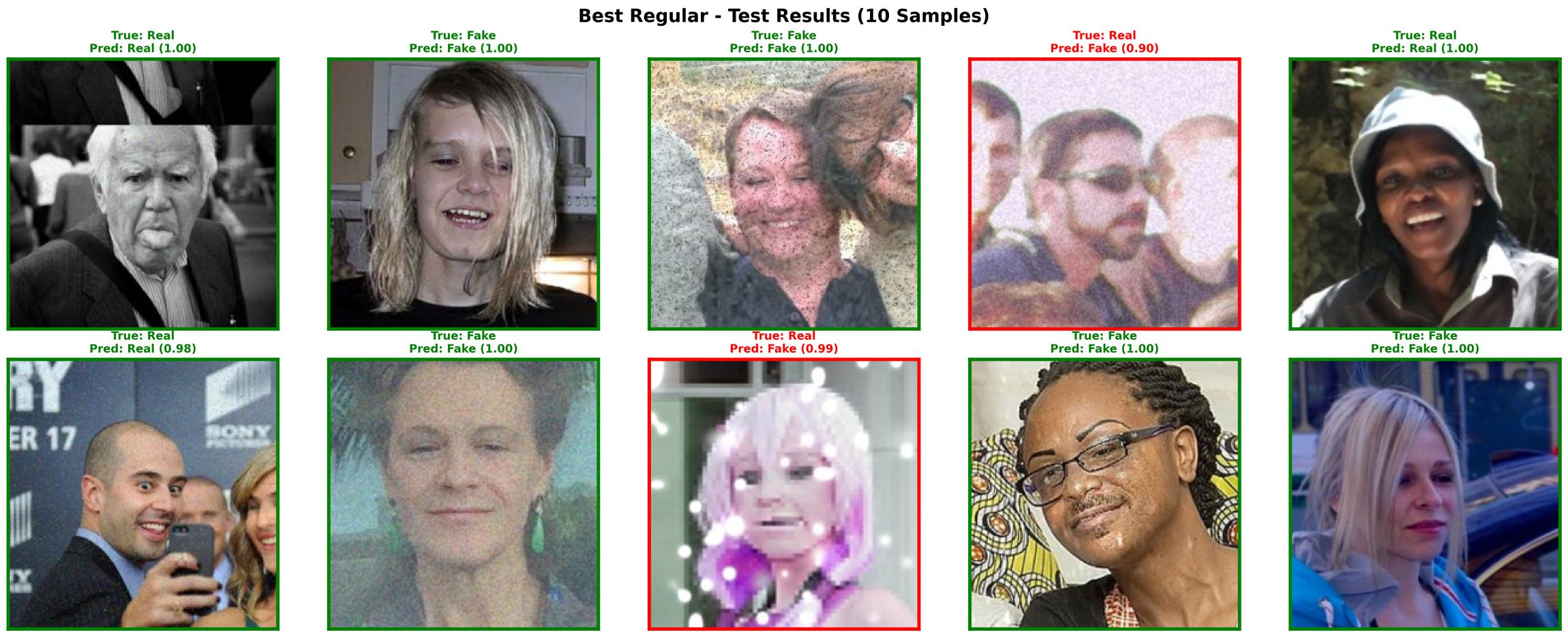}\\[2pt]
\includegraphics[width=0.73\textwidth]{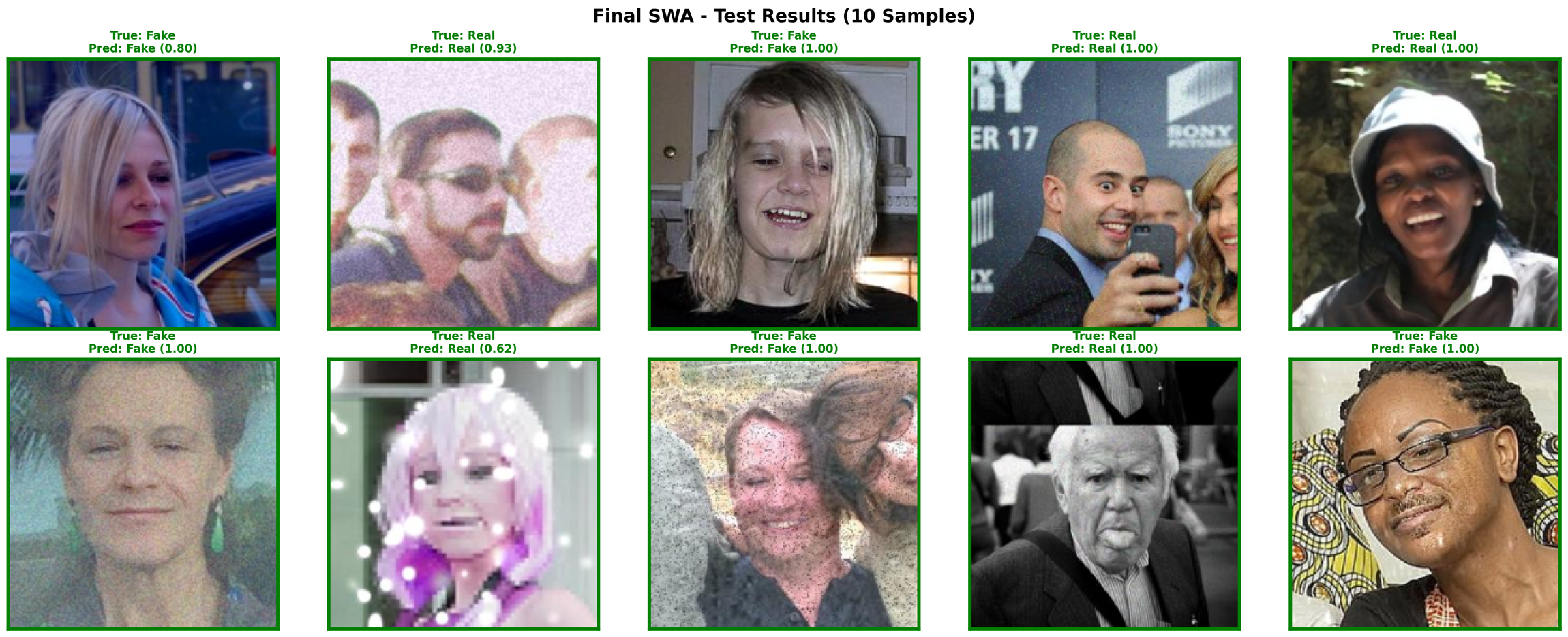}
\caption{Ten randomly drawn OpenForensics test images under the
validation-selected checkpoint (top) and the terminal average (bottom); same
images, different order. Titles give label, prediction and softmax confidence;
red borders mark errors. Both errors in the top panel are degraded genuine
photographs --- a heavily noised crowd scene ($0.90$) and a portrait with an
extreme magenta cast ($0.99$) --- rejected as fake. The average classifies both
correctly, with its two lowest confidences ($0.93$ and $0.62$).}
\label{fig:qual}
\end{figure*}

\subsection{Comparison on DF-Wild-Cup}\label{sec:spcup}
PairCLIP-SWA improves on CAE-Net on every metric both systems report
(Tables~\ref{tab:spcup} and~\ref{tab:head2head}). At the
default threshold, accuracy rises from $94.46$ to $97.04\%$, a $46.5\%$ reduction
in error ($170\to91$ misclassified images), and the two Wilson intervals do
not overlap. The threshold-free metrics, which do not depend on any operating
point, show the same ordering: AU-ROC $98.64$ vs.\ $97.60\%$ ($43\%$ less area
above the curve) and EER $4.07$ vs.\ $5.00\%$. The margin of $2.58$ points is
$4.4$ times the seed standard deviation CAE-Net reports for its final ensemble
($\pm0.58$); we cannot report our own seed dispersion
(Section~\ref{sec:limits}). Even the non-averaged checkpoint ($95.67\%$) exceeds
the ensemble.

At $\tau^\star$ the operating point is strongly asymmetric: specificity is
$99.03\%$ and fake-class recall $95.54\%$, so $68$ of the $83$ remaining errors
are forgeries accepted as real. Moving the threshold on $P(\text{real})$ from
$0.5$ to $0.680$ gains only $0.26$ points of accuracy, and because $\tau^\star$
is chosen on the evaluation split, that gain is an upper bound.

The cost side is explicit. The detector is $2.59\times$ larger, needs
$3.27\times$ the MACs, and measured $2.7\times$ the per-image latency of
CAE-Net, although on different hardware. In exchange it uses one backbone, one
forward pass, one training stage and a third of the training images, with no
wavelet branch and no fusion weights to tune (Fig.~\ref{fig:tradeoff}).

\begin{figure*}[!t]
\centering
\includegraphics[width=\textwidth]{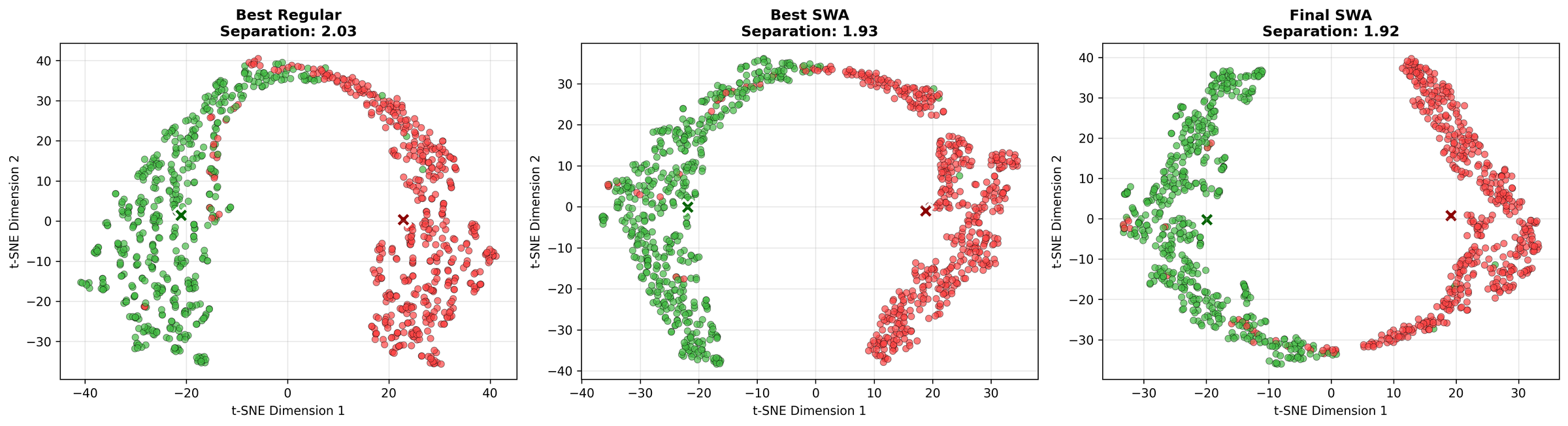}
\caption{t-SNE embeddings of $1{,}000$ DF-Wild-Cup validation images ($493$
fake, $507$ real) for the three checkpoints; green is real, red fake, crosses
are centroids.}
\label{fig:tsne-compare}
\end{figure*}

\begin{table}[!tb]
\caption{Geometry and k-NN probe in t-SNE space. DF-Wild-Cup: $1{,}000$
validation images; OpenForensics: $1{,}000$ held-out images ($513$ fake,
$487$ real). ``Head'' is classifier accuracy on the full split.}
\label{tab:geometry}
\centering
\footnotesize
\setlength{\tabcolsep}{3.2pt}
\begin{tabular}{@{}llrrrrrr@{}}
\toprule
Corpus & Checkpoint & $\Delta$ & $\sigma_f$ & $\sigma_r$ & $S$ & Probe & Head \\
\midrule
\multirow{3}{*}{DF-Wild-Cup}
 & Best regular & $43.90$ & $10.70$ & $10.90$ & $\best{2.033}$ & $96.00$ & $95.67$ \\
 & Best SWA     & $40.73$ & $10.41$ & $10.72$ & $1.927$ & $\best{97.20}$ & $\best{97.04}$ \\
 & Final SWA    & $39.08$ & $10.32$ & $10.07$ & $1.917$ & $\best{97.20}$ & $\best{97.04}$ \\
\midrule
\multirow{3}{*}{\makecell[l]{Open-\\Forensics}}
 & Best regular & $40.42$ & $8.42$ & $9.72$ & $2.228$ & $97.80$ & $88.85^\ddagger$ \\
 & Best SWA     & $40.42$ & $8.42$ & $9.72$ & $2.228$ & $97.80$ & $88.85^\ddagger$ \\
 & Final SWA    & $43.02$ & $8.97$ & $10.09$ & $\best{2.257}$ & $97.80$ & $\best{94.76}^\ddagger$ \\
\bottomrule
\multicolumn{8}{@{}l}{\scriptsize $^\ddagger$Test partition.}
\end{tabular}
\end{table}

\subsection{OpenForensics: Convergence and Validation}\label{sec:val}
The OpenForensics run converges quickly: at epoch $11$ training and in-loop
validation accuracy are $98.46$ and $98.64\%$, with a validation loss of
$0.0426$; re-evaluated in full precision, the best-regular and best-SWA
checkpoints score $98.43\%$ ($38{,}809/39{,}428$) and the final average
$97.67\%$ ($38{,}509/39{,}428$). Under the
schedule of Eq.~\eqref{eq:schedule} at most seven of the $24$ blocks are
trainable by that epoch, and the validation accuracy slightly exceeds the
training accuracy, so there is no sign of memorization on the training
distribution. A small train--validation gap says nothing, however, about how the
selected checkpoint behaves on a split it was not selected on.

On validation the selected checkpoint is nearly symmetric (fake recall
$99.03\%$, specificity $97.83\%$, MCC $0.969$) and beats the terminal average,
which trades recall for specificity ($95.73$ vs.\ $99.60\%$). Read in isolation,
this split recommends the best-regular checkpoint without hesitation.

\begin{table}[!tb]
\caption{Ablations on the DF-Wild-Cup validation split ($n=3{,}072$). Rows marked
\yes{} differ from the reported configuration only in the stated factor; the
others also differ in schedule or backbone and bound, rather than isolate, the
effect. Train accuracy shows the memorization gap where logged.}
\label{tab:ablation}
\centering
\footnotesize
\setlength{\tabcolsep}{3pt}
\begin{tabular}{@{}lrrc@{}}
\toprule
Variant & Train & Val. & Ctrl. \\
\midrule
\multicolumn{4}{@{}l}{\emph{A. Input representation (P1)}} \\
Full released frame (reported run) & $98.48$ & $\best{97.04}$ & ref. \\
Face crop (dlib~\cite{king2009dlib}) $+$ OpenCLIP & $84.34$ & $56.05$ & --- \\
FFT-domain input & $83.72$ & $49.61$ & --- \\
Wavelet-coefficient input, 30 ep & $91.60$ & $73.20$ & --- \\
\midrule
\multicolumn{4}{@{}l}{\emph{B. Class balancing (P6), OpenCLIP ViT-L/14}} \\
Similarity-guided $1{:}1$, $k{\to}11$, non-avg. & --- & $95.67$ & ref. \\
Random $1{:}1$, $k$: $2{\to}5$ & --- & $93.36$ & --- \\
\midrule
\multicolumn{4}{@{}l}{\emph{C. Augmentation (P5), ViT $+$ LoRA~\cite{hu2022lora}}} \\
None, 3 ep & --- & $86.75$ & --- \\
Augmented, 10 ep: peak / final & --- & $91.44$/$87.99$ & --- \\
\midrule
\multicolumn{4}{@{}l}{\emph{D. Training objective}} \\
Cross-entropy (reported run) & $98.48$ & $\best{97.04}$ & ref. \\
Triplet loss~\cite{schroff2015facenet} $+$ classifier & $98.70$ & $90.85$ & --- \\
\midrule
\multicolumn{4}{@{}l}{\emph{E. Same run, single factor}} \\
SWA vs.\ non-averaged & --- & $97.04$/$95.67$ & \yes \\
$\tau^\star$ vs.\ $0.5$ (tuned on split) & --- & $97.30$/$97.04$ & \yes \\
Before adaptation (random head) & --- & $45.80$ & \yes \\
\bottomrule
\end{tabular}
\end{table}

\begin{figure*}[!t]
\centering
\includegraphics[width=\textwidth]{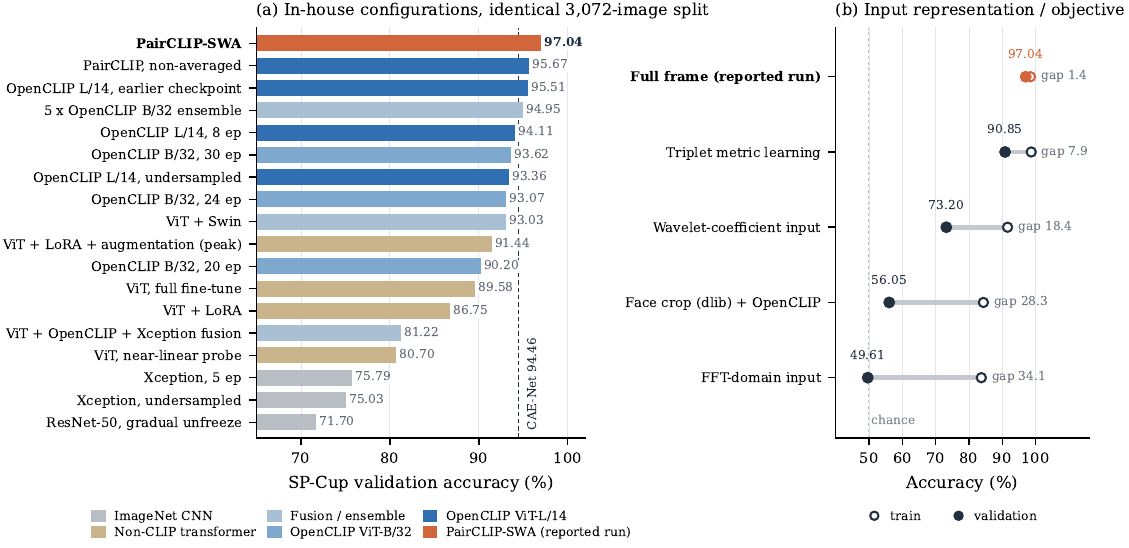}
\caption{(a) In-house configurations evaluated on the identical
DF-Wild-Cup validation split, coloured by backbone family; the dashed line is
CAE-Net. (b) Train (open) and validation (filled) accuracy for alternative input
representations and objectives; the gap is the memorization that each fails to
convert into validation accuracy.}
\label{fig:design}
\end{figure*}

\subsection{OpenForensics: the Held-Out Test Partition}\label{sec:test}
The test partition is drawn from the same corpus as the validation partition,
so the shift it measures is the residual difference between two disjoint
samples of in-the-wild imagery --- the kind of shift that separates any
benchmark split from deployment data~\cite{recht2019imagenet}. It is enough to
reverse the ranking (Fig.~\ref{fig:valtest}, Table~\ref{tab:ofmetrics}).

The validation-selected checkpoints fall to $88.85\%$, $9.58$ points below
their validation accuracy. The loss is one-sided: fake-class recall barely
moves ($99.03\to98.67\%$) while specificity collapses ($97.83\to78.88\%$).
Of $1{,}216$ test errors, $1{,}143$ ($94.0\%$) are genuine photographs rejected
as forgeries, a $15.7{:}1$ asymmetry. The terminal average, ranked last on
validation, reaches $94.76\%$: false positives fall from $1{,}143$ to $275$,
false negatives rise from $73$ to $296$, the asymmetry becomes $0.93{:}1$, MCC
rises from $0.792$ to $0.895$, and test error falls by $53.0\%$. The Wilson
intervals, $[88.24, 89.43]$ and $[94.33, 95.17]$, are more than five points
apart. Selecting by validation accuracy would have deployed the model that
makes $2.1\times$ as many errors.

Fig.~\ref{fig:qual} shows the mechanism in miniature. The sample spans sensor
noise, low light, greyscale, motion blur, JPEG blocking, colour casts, occlusion
and multi-face crops. The two images the selected checkpoint misclassifies are
both genuine and both severely degraded, and it rejects them with high
confidence. The average gets both right, but with low confidence. This is
consistent with the selected checkpoint having learned that low-level
degradation indicates synthesis --- an association that preprocessing step~P5,
which never shows the network degraded real images, does nothing to discourage
(Section~\ref{sec:discussion}). Ten images illustrate; the $10{,}905$ quantify.

\begin{table}[!tb]
\caption{In-house configurations on the DF-Wild-Cup validation split, grouped by
backbone. Data: S = SP-Cup training partition, P = Stage-1 subsample,
U = random $1{:}1$ undersample. ``Init.'' is accuracy with an untrained head.
Schedules differ between rows.}
\label{tab:ladder}
\centering
\footnotesize
\setlength{\tabcolsep}{3pt}
\begin{tabular}{@{}lcrr@{}}
\toprule
Configuration & Data & Init. & Val. \\
\midrule
\multicolumn{4}{@{}l}{\emph{ImageNet-initialized CNNs}} \\
ResNet-50, gradual unfreezing, 40 ep & P & --- & $71.70$ \\
Xception, 5 ep (two runs) & S & --- & $75.16$, $75.79$ \\
Xception, undersampled to $85{,}380$ & U & --- & $75.03$ \\
\midrule
\multicolumn{4}{@{}l}{\emph{Non-CLIP transformers}} \\
ViT, $0.69\%$ unfrozen (probe), 30 ep & P & $53.74$ & $80.70$ \\
ViT $+$ LoRA, 3 ep & S & --- & $86.75$ \\
ViT, full fine-tuning, 2 ep (ends $87.60$) & S & --- & $89.58$ \\
ViT $+$ LoRA $+$ aug., 10 ep (ends $87.99$) & S & --- & $91.44$ \\
ViT $+$ Swin, 10 ep & S & --- & $93.03$ \\
\midrule
\multicolumn{4}{@{}l}{\emph{Fusion and prediction ensembles}} \\
ViT $+$ OpenCLIP $+$ Xception fusion, 8 ep & S & --- & $81.22$ \\
$5\times$ B/32, disjoint parts (MV $94.66$) & P & --- & $94.95$ \\
\midrule
\multicolumn{4}{@{}l}{\emph{OpenCLIP ViT-B/32}} \\
20 ep & P & $51.76$ & $90.20$ \\
24 ep & P & $50.10$ & $93.07$ \\
30 ep & P & $49.80$ & $93.62$ \\
\midrule
\multicolumn{4}{@{}l}{\emph{OpenCLIP ViT-L/14}} \\
8 ep & P & $45.61$ & $94.11$ \\
Unfreezing $2\to5$ blocks & U & --- & $93.36$ \\
Earlier checkpoint, re-evaluated & --- & --- & $95.51$ \\
$k$: $2{\to}11$, non-averaged (reported) & P & $45.80$ & $95.67$ \\
$k$: $2{\to}11$ $+$ SWA (reported) & P & $45.80$ & $\best{97.04}$ \\
\bottomrule
\end{tabular}
\end{table}

\subsection{Representation Geometry}\label{sec:cluster}
Table~\ref{tab:geometry} and Fig.~\ref{fig:tsne-compare} support one
conclusion from two directions.

\emph{On DF-Wild-Cup, separation and accuracy move in opposite directions.}
The best-regular checkpoint has the highest separation score ($S=2.033$) and
the lowest accuracy. Averaging shrinks $\Delta$ by $11.0\%$ but the pooled
dispersion by only $5.6\%$, so $S$ falls to $1.917$, while the probe removes
$14$ of its $35$ fake-side errors (fake recall $92.90\to95.74\%$, MCC
$0.922\to0.944$) and the head gains $1.37$ points.

\emph{On OpenForensics, near-identical geometry hides a six-point gap.} The
checkpoints differ by at most $0.03$ in $S$ and not at all in probe accuracy
($97.80\%$), yet by $5.91$ points on the held-out test partition. What averaging
changes is not the arrangement of the embedding but the decision boundary laid
over it --- precisely the quantity a separability index does not measure.

\emph{Without projection, cluster structure is weak.} DBSCAN on
$768$-dimensional OpenCLIP features gives silhouette coefficients of $0.109$
and $0.315$ with Davies--Bouldin indices of $0.911$ and $0.700$ for its two
parameter settings --- moderate unsupervised structure at best, in a feature
family on which a linear head exceeds $97\%$. Cluster indices, projected or not,
are not proxies for classifier quality, and reporting them as evidence of
representation quality, as is common (e.g.,~\cite{bhattacharjee2026caenet}),
warrants that caveat.
The DF-Wild-Cup probe agrees with the head to within $0.16$ points, but it is
fit on a single $1{,}000$-image subsample, and the differences in $S$ are small
enough that subsample variance cannot be excluded (Section~\ref{sec:limits}).

\section{Ablations, Design-Space Study and Discussion}\label{sec:discussion}

\subsection{Preprocessing and Data-Preparation Ablations}\label{sec:ablation}
Table~\ref{tab:ablation} and Fig.~\ref{fig:design}(b) collect every
preprocessing and data-preparation alternative we evaluated on the identical
split. Only block~E is controlled; the others are honest bounds.

\emph{Secondary face cropping is the most damaging choice we tested.} Cropping
to a dlib face box before an OpenCLIP classifier yields $56.05\%$, about $34$
points below the same model on full frames, while still fitting $84.34\%$ of the
training data. The $28$-point memorization gap is the signature of a model
that has lost access to transferable evidence and has fitted something
corpus-specific instead. This is the empirical basis for P1: DF-Wild-Cup images
are already aligned face regions, and the band of context around the face ---
where blending boundaries~\cite{li2020facexray} live --- is worth more than the
extra resolution a tighter crop buys.

\emph{Frequency-domain inputs transferred poorly.} A classifier on FFT spectra
fits $83.72\%$ of the training data and scores $49.61\%$ --- chance --- on
validation; a wavelet-coefficient input reaches $73.20\%$, $16$ points below
the ConvNeXt$+$wavelet result reported in~\cite{bhattacharjee2026caenet}. That
result shows wavelet inputs can work with a suitable backbone, but even it
remains $7.55$ points below spatial RGB with CLIP initialization. We therefore
feed the network spatial RGB (P2--P4) and leave spectral sensitivity to the
pre-trained filters.

\emph{Balancing strategy.} With the same OpenCLIP ViT-L/14 backbone, the
similarity-guided subsample reaches $95.67\%$ (non-averaged) against $93.36\%$
for a random one-to-one undersample. The runs also differ in unfreezing depth
($11$ vs.\ $5$ blocks) and, given the Stage-1 preprocessing issue of
Section~\ref{sec:repro}, the $2.31$-point difference cannot be attributed to
semantic content matching. It is consistent with, but does not establish, a
benefit from Stage~1.

\emph{Augmentation.} For a ViT with LoRA adapters, augmentation raised the
early validation peak from $86.75$ to $91.44\%$ but the model ended at
$87.99\%$ after ten epochs, retaining little of the gain. Together with the
forensic argument for P5, this led us to flip-only training. The OpenForensics
error analysis suggests where that choice may fall short: a detector never shown
degraded real images can learn that degradation implies forgery. The principled
remedy is \emph{class-symmetric} corruption --- applying the same
noise, blur and compression~\cite{hendrycks2019corruptions} to both classes, so
degradation carries no label information --- which we have not yet tested.

\emph{Objective.} Triplet metric learning reaches the highest training accuracy
of any variant ($98.70\%$) and a validation accuracy $6.19$ points below
cross-entropy: a pair-based alternative to Stage~1 that memorizes rather than
generalizes.

\subsection{Backbone and Adaptation: a Design-Space Study}\label{sec:ladder}
Table~\ref{tab:ladder} and Fig.~\ref{fig:design}(a) record twenty-two
configurations on the identical split, counting the four of
Table~\ref{tab:ablation} not repeated here. Their schedules, epochs and in some
cases training data differ, so the table is a survey of the design space rather
than a controlled ablation, and single differences should not be over-read.
Four regularities are nevertheless robust to those confounds.

\emph{Pre-training dominates.} Grouped by initialization, ImageNet CNNs span
$71.70$--$75.79\%$, non-CLIP transformers $80.70$--$93.03\%$, OpenCLIP ViT-B/32
$90.20$--$93.62\%$ and OpenCLIP ViT-L/14 $93.36$--$97.04\%$. Every untrained head
starts at chance ($45.61$--$53.74\%$), so the ordering reflects adaptation, not
leakage. CLIP initialization is worth more than any architectural or schedule
change within a family --- the empirical form of gap G2.

\emph{Adapting too little or too much both lose.} A near-linear probe on a
non-CLIP ViT reaches $80.70\%$; full fine-tuning reaches $89.58\%$ after one
epoch and degrades to $87.60\%$ after two; adapter-based tuning with LoRA sits
between. Progressive unfreezing of an OpenCLIP tower is the best-performing
strategy we tried, consistent with~\cite{kumar2022finetuning}.

\emph{Averaging weights beats averaging predictions.} Five OpenCLIP ViT-B/32
models trained on disjoint data parts average $92.75\%$ individually and reach
$94.95\%$ as an ensemble ($94.66\%$ by majority vote), at five forward passes. A
single ViT-L/14 with SWA reaches $97.04\%$ in one pass, although at higher
per-pass cost. Feature-level fusion of three heterogeneous backbones
($81.22\%$) falls below every single-backbone transformer except the
near-linear probe.

\emph{Scale within CLIP helps, but schedule helps more.} OpenCLIP ViT-L/14 after
eight epochs ($94.11\%$) is only $0.49$ points above the best ViT-B/32; the
reported ViT-L/14 run gains a further $1.56$ points with a longer schedule and
deeper progressive unfreezing, and $1.37$ from averaging.

\subsection{What Weight Averaging Buys, and Why}
The two selection splits disagree about averaging ($+1.37$ on DF-Wild-Cup,
$-0.76$ on OpenForensics), and a single mechanism accounts for both. The
non-averaged iterate is a high-variance object: on DF-Wild-Cup its validation
accuracy has a standard deviation of $2.13$ points over eight epochs, its
validation loss doubles while its training loss falls, and its best epoch is a
maximum over noise. Averaging integrates over that noise.

When the iterate's noise happens to align with the selection split, as on
OpenForensics validation, the selected iterate beats the average by a small
margin; when it does not, as throughout the DF-Wild-Cup trajectory, the average
wins at every epoch. On data used for neither training nor selection, the
selected iterate's advantage is not merely lost but reversed, by $5.91$ points.
This is the pattern that selection bias predicts~\cite{cawley2010overfitting}:
the more variable the candidates, the more the maximum over them overstates the
performance of the chosen one. Weight averaging reduces that variance at zero
inference cost, and the flat-minimum account of SWA~\cite{izmailov2018swa}
explains why the averaged solution degrades less under shift.

\subsection{Validation Accuracy Is a Poor Deployment Proxy}
The OpenForensics result is the most transferable lesson in this paper. The
standard protocol --- select the checkpoint with the best validation accuracy
--- chose a model that makes $2.1\times$ as many errors on held-out data, and the
failure was invisible to it for a structural reason: the loss was confined to
one class, and scalar accuracy on a balanced split weights a specificity
collapse no more heavily than a symmetric error. We therefore recommend that
forensic detectors be (i) selected on balanced accuracy or MCC, (ii) on a
partition disjoint from the one used for early stopping, with (iii) per-class
recall reported alongside accuracy, and (iv) weight averaging preferred to a
single selected iterate. The same caution applies to every DF-Wild-Cup number
in Table~\ref{tab:spcup}, ours included: each is measured on the split its
system was selected or tuned on, and should be read as an upper bound on
deployment accuracy.

\subsection{Computational Complexity}
The forward cost is dominated by the $24$ blocks. For $N=N_p+1$ tokens of width
$d_m$, a block costs $4Nd_m^2$ MACs in its projections, $2N^2d_m$ in attention
and $8Nd_m^2$ in its MLP:
\begin{equation}
\mathrm{MACs}=L\big(12Nd_m^2+2N^2d_m\big)+N_p\cdot3\cdot14^2\cdot d_m .
\label{eq:macs}
\end{equation}
With $L=24$, $N=257$, $d_m=1024$, this is $81.0$ GMACs ($162.0$ GFLOPs) per
image; the quadratic attention term contributes $4.0\%$. Parameter memory is
$1{,}160$\,MB in \texttt{float32} and $580$\,MB in \texttt{float16}, against
$448$\,MB for CAE-Net; measured latency was $28.1$\,ms per image on
$2\times$T4. Training is cheaper than these figures suggest, because between
$8.3\%$ and $45.6\%$ of parameters received gradients in the reported run, and
one stage on $85{,}380$ images replaces five stages on $86{,}584$ each. Stage~1
costs $n_r+n_f$ frozen forward passes plus $9.37\times10^9$ similarities, once.

\subsection{On the Absence of Weight Decay}
With $\lambda=0$, regularization is structural: $54.42\%$ of the tower never
moves in the reported run, batches are exactly balanced, and averaging smooths
the trajectory. An explicit penalty would pull the trainable suffix away from
its pre-trained values, the distortion~\cite{kumar2022finetuning} associates
with lost robustness. The trajectory of Fig.~\ref{fig:trajectory}(b) shows,
however, that the non-averaged suffix does overfit, so a sweep over $\lambda$
remains worth running.

\subsection{Limitations}\label{sec:limits}
\begin{enumerate}
\item \textbf{Single seed.} The reported runs are single trajectories. CAE-Net
reports five-seed means; our $2.58$-point margin is $4.4\times$ its seed
deviation, but our own dispersion is unknown.
\item \textbf{Stage~1 is not isolated, and its reported implementation
double-rescaled its inputs.} No control with a random one-to-one subsample
under an identical schedule exists, and the similarities used for the reported
corpus were computed on attenuated inputs (Section~\ref{sec:repro}).
\item \textbf{Selection on the evaluation split.} DF-Wild-Cup numbers,
including CAE-Net's, are selection-split numbers; $\tau^\star$ is also tuned
there.
\item \textbf{Uncontrolled design-space runs.} Apart from block~E of
Table~\ref{tab:ablation}, comparisons in Section~\ref{sec:discussion} confound
schedule, epochs and, in places, training data and hardware.
\item \textbf{Metric coverage.} Threshold-free metrics exist only for the
DF-Wild-Cup best-SWA checkpoint; the OpenForensics analysis is at
$\tau=0.5$.
\item \textbf{Geometry from one subsample.} t-SNE, separation scores and
probes are computed on one $1{,}000$-image subsample per corpus.
\item \textbf{No adversarial evaluation}, unlike~\cite{bhattacharjee2026caenet}.
\item \textbf{Cost.} $304$\,M parameters and $81$ GMACs rule out edge
deployment without distillation.
\end{enumerate}

\section{Conclusion and Future Work}\label{sec:conclusion}
We set out to build a single-backbone deepfake detector and found that the way
its checkpoints are selected matters as much as the detector. On DF-Wild-Cup,
PairCLIP-SWA --- an OpenCLIP ViT-L/14 tower adapted by progressive unfreezing on a
one-to-one balanced subsample and averaged in weight space --- improves on the
best published three-backbone ensemble in accuracy ($97.04$ vs.\ $94.46\%$),
AU-ROC ($98.64$ vs.\ $97.60\%$), EER ($4.07$ vs.\ $5.00\%$) and $F_1$, under the
same in-domain protocol and with a third of the training images. Its training
log shows the non-averaged checkpoint swinging across $5.34$ points while the
average is never worse. On an OpenForensics-derived benchmark, the checkpoint
chosen by validation accuracy made $2.1\times$ as many errors on a held-out
partition as the checkpoint ranked last, through a one-sided specificity
collapse that scalar accuracy could not see. Twenty-two configurations on the
identical split show that CLIP pre-training, not ensembling, drives accuracy,
that secondary face cropping and frequency-domain inputs discard transferable
evidence, and that cluster separability does not track decision quality.

The next experiments follow directly from the limitations. (i) Re-run Stage~1
with corrected preprocessing and train a random-subsample control under an
identical schedule, to measure what similarity-guided balancing contributes.
(ii) Repeat the main run with additional seeds, and re-run the threshold sweep
on the OpenForensics splits, so that selection effects carry dispersion and
threshold-free metrics. (iii) Evaluate FGSM~\cite{goodfellow2015fgsm} and
PGD~\cite{madry2018pgd} robustness. (iv) Test class-symmetric corruption
augmentation against the degradation shortcut identified in
Section~\ref{sec:test}. (v) Distil the averaged tower into a smaller student and
measure how much of its accuracy and its selection robustness survive.

\section*{Acknowledgment}
The authors thank the IEEE Signal Processing Society for organizing the Signal
Processing Cup 2025 and releasing the DF-Wild-Cup dataset, and the maintainers
of OpenCLIP for the pre-trained models used in this study.

\bibliographystyle{IEEEtran}
\bibliography{references}

\end{document}